\documentclass[11pt]{article}

\usepackage[final]{acl}

\usepackage{times}
\usepackage{latexsym}

\usepackage[most]{tcolorbox}
\usepackage{listings}
\usepackage[table]{xcolor}
\usepackage{bm}

\usepackage[T1]{fontenc}
\usepackage[utf8]{inputenc}

\usepackage{microtype}
\usepackage{inconsolata}

\usepackage{graphicx}
\usepackage{pifont}     
\usepackage{xcolor}     
\usepackage{multirow}
\usepackage{subcaption}
\usepackage{amsmath}
\usepackage{amssymb}
\usepackage{amsfonts}
\usepackage{siunitx}
\usepackage{booktabs}

\title{SGG-R\textsuperscript{3}: From Next-Token Prediction to End-to-End Unbiased Scene Graph Generation}

\author{
  \textbf{Jiaye Feng}\textsuperscript{1}, 
  \textbf{Qixiang Yin}\textsuperscript{2}, 
  \textbf{Yuankun Liu}\textsuperscript{1}, 
  \textbf{Tong Mo}\textsuperscript{1*}, 
  \textbf{Weiping Li}\textsuperscript{1} \\
  \textsuperscript{1}School of Software and Microelectronics, Peking University, Beijing, China \\
  \textsuperscript{2}Zhongguancun Academy, Beijing, China \\
  \texttt{\{jyfeng25, liuyk\}@stu.pku.edu.cn}, \\
  \texttt{buptyqx@bupt.edu.cn}, \texttt{\{motong, wpli\}@ss.pku.edu.cn} \\
  \small \textsuperscript{*}Corresponding author
}

\begin{document}
\maketitle

\begin{abstract}
Scene Graph Generation (SGG) structures visual scenes as graphs of objects and their relations. While Multimodal Large Language Models (MLLMs) have advanced end-to-end SGG, current methods are hindered by both a lack of task-specific structured reasoning and the challenges of sparse, long-tailed relation distributions, resulting in incomplete scene graphs characterized by low recall and biased predictions. To address these issues, we introduce SGG-R\textsuperscript{3}, a structured reasoning framework that integrates task-specific Chain-of-Thought (CoT)-guided Supervised Fine-Tuning (SFT) and Reinforcement Learning (RL) with Group Sequence Policy Optimization (GSPO), designed to engage in three sequential stages to achieve end-to-end unbiased scene graph generation. During the SFT phase, we propose a relation augmentation strategy by leveraging an MLLM and refined via embedding similarity filtering to alleviate relation sparsity. Subsequently, a stage-aligned reward scheme optimizes the procedural reasoning during RL. Specifically, we propose a novel dual-granularity reward which integrates fine-grained and coarse-grained relation rewards, simultaneously mitigating the long-tail issue via frequency-based adaptive weighting of predicates and improving relation coverage through semantic clustering. Experiments on two benchmarks show that SGG-R\textsuperscript{3} achieves superior performance compared to existing methods, demonstrating the effectiveness and generalization of the framework.
\end{abstract}

\begin{figure}[t]
  \includegraphics[width=1.0\linewidth]{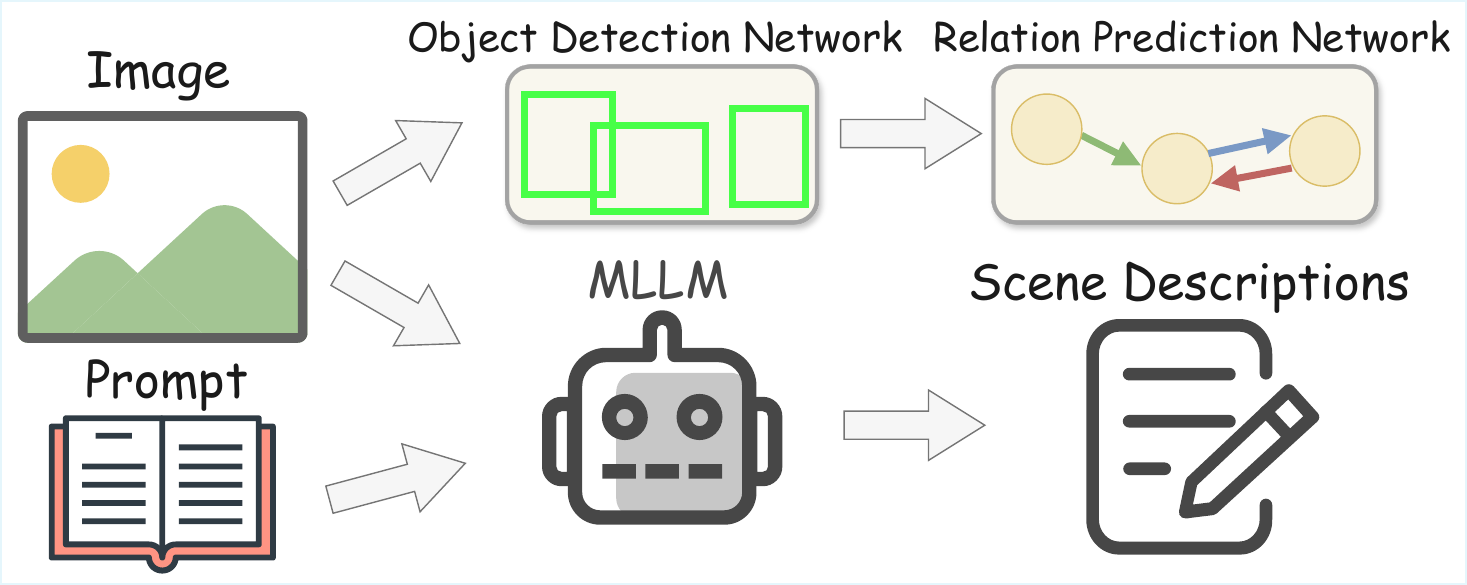}
  \caption{Pipeline comparison of Scene Graph Generation between the two-stage classification framework and the end-to-end generative MLLM method.}
  \label{fig:sgg1}
\end{figure}

\section{Introduction}

Scene Graph Generation (SGG) aims to parse images into structured semantic graphs that represent objects and their interrelationships, providing a crucial bridge between vision and language that facilitates downstream applications including visual question answering~\cite{kenfack2020robotvqa}, image retrieval~\cite{schroeder2020structured}, and embodied navigation~\cite{zhu2021hierarchical, gadre2022continuous, huang2025esca}. Traditional SGG methods~\cite{xu2017scene, zellers2018neural, tang2019learning} predominantly follow a two-stage paradigm, decoupling the task into object detection followed by relation prediction, with efforts concentrated on refining the latter module. To improve efficiency and mitigate error propagation, end-to-end SGG approaches~\cite{lin2020gps, li2022sgtr, cong2023reltr, hayder2024dsgg} have emerged, which are often based on Transformer~\cite{vaswani2017attention} architectures that unify object extraction and relation analysis. However, such methods tend to produce predictions that are biased toward high-frequency relations and have limited generalization to unseen concepts due to overfitting on annotated data.

Recently, Multimodal Large Language Models (MLLMs)~\cite{liu2023visual, wang2024qwen2, bai2025qwen2} have advanced rapidly, demonstrating considerable capabilities in object grounding and relational reasoning. Unlike conventional classification-based SGG methods that treat relation prediction as discrete labeling over fixed object proposals, MLLMs facilitate the end-to-end generation of scene descriptions in an open-ended manner. As illustrated in Figure~\ref{fig:sgg1}, guided by the task prompt, the model directly generates a textual representation of the scene graph, eliminating the need for the two-stage generation process required by conventional methods. However, applying MLLM to end-to-end SGG presents two critical challenges. First, the absence of SGG-specific structured reasoning pipeline compels model to navigate an excessively vast search space. For instance, using generic directives like ``think step-by-step in \texttt{<think></think>} tags'' without granular procedural guidance makes it difficult to form a valid Chain-of-Thought (CoT), leading to hallucinations and low recall. Second, the inherent sparsity and long-tailed distribution of relation data (detailed analysis in Appendix~\ref{app:Dataset Analysis}) drive fine-tuned models to predict incomplete graphs with biased relations. While existing research~\cite{chen2025compile, li2025relation} utilizes Supervised Fine-Tuning (SFT) and Reinforcement Learning (RL) to optimize output, these efforts often struggle with precise object grounding and unbiased relational modeling due to the absence of a specialized structured reasoning process.

Notably, SGG fundamentally differs from typical visual question answering, as it requires dense and sequential prediction of both objects and their relations, thus necessitating a tailored CoT process. To this end, we introduce SGG-R\textsuperscript{3}, a structured reasoning framework that integrates CoT-guided SFT and RL paradigm. To alleviate relation sparsity and the long-tailed distribution issue, we first generate relation-augmented data by leveraging Qwen2.5-VL-32B~\cite{bai2025qwen2}. This dataset is then filtered by computing the cosine similarity between Sentence-BERT~\cite{reimers2019sentence} embeddings of the generated relations and the original dataset relations, thereby maintaining semantic alignment and yielding higher-quality samples for SFT. To establish foundational reasoning capabilities, the model is supervised fine-tuned on relation-augmented data formatted as CoT, explicitly engaging in three sequential steps of object category detection, object instance grounding, and multi-type relation extraction. The RL phase further enhances the model's capabilities in visual grounding and relational reasoning. We introduce a novel dual-granularity reward that integrates fine-grained and coarse-grained relation rewards, which adaptively weights predicates based on their frequency to alleviate the long-tail issue, while also enhancing overall relation coverage by using DBSCAN-based~\cite{ester1996density} semantic clustering to group and match similar relations. Our main contributions are summarized as follows:

\begin{itemize}
    \item We propose a structured reasoning framework for end-to-end unbiased scene graph generation, which enhances controllability and logical coherence by decomposing the generation into three sequential stages.
    
    \item We introduce a relation augmentation method for SFT and a novel dual-granularity reward for RL, jointly mitigating the long-tailed distribution issue and promoting the generation of unbiased scene graphs.
    
    \item Extensive experiments on two benchmarks show that SGG-R\textsuperscript{3} outperforms previous methods across multiple metrics, demonstrating its superior capability in visual grounding and relation identification.
\end{itemize}

\begin{figure*}[h]
  \centering
  \includegraphics[width=\textwidth]{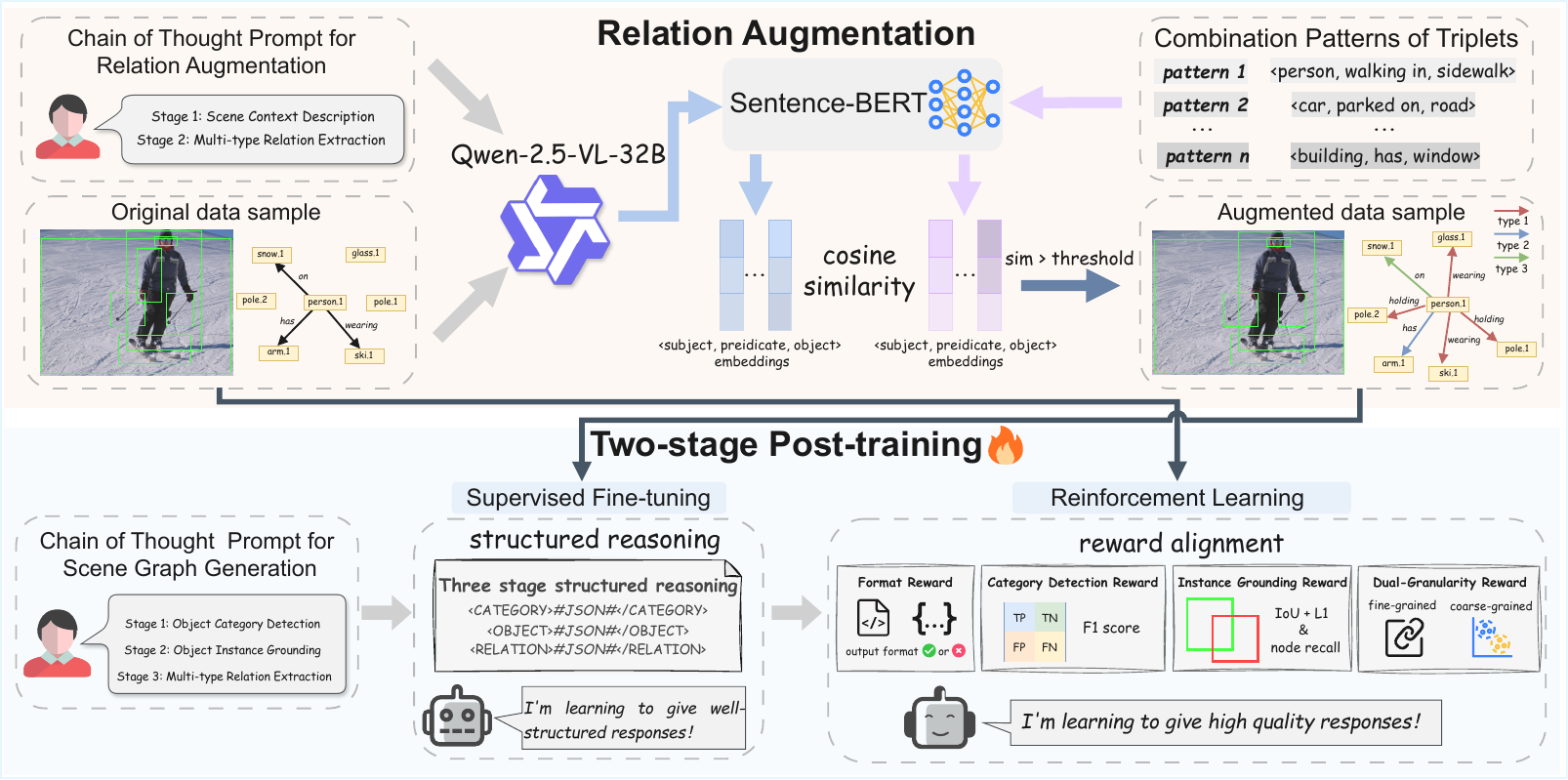}  
  \caption{Overview of the SGG-R\textsuperscript{3} framework. The ``R\textsuperscript{3}'' denotes three key contributions: \textbf{R}elation augmentation, structured \textbf{R}easoning, and \textbf{R}eward alignment. First, candidate relations generated by the Qwen2.5-VL-32B with CoT prompt are filtered via Sentence-BERT embedding similarity against the original data. The model is then supervised fine-tuned on the CoT-formatted augmented data with CoT prompt, followed by reward-driven reinforcement learning aligned with the original dataset.}
  \label{fig:framework}
\end{figure*}

\section{Related Works}
\subsection{VLM for Scene Graph Generation}
Vision-Language Models (VLMs) provide a promising approach for SGG by leveraging cross-modal alignment. Current works~\cite{zhang2023learning, li2024pixels, chen2024expanding} typically retrieve objects and relations through global or regional feature matching for open-vocabulary SGG. \cite{yu2023visually} tackles the long-tailed distribution by mining fine-grained predicates via a VLM and an adaptive semantic clustering loss. Meanwhile, MLLMs are fine-tuned to generate scene graphs in an end-to-end manner via SFT and RL. \cite{chen2025compile} pioneers the application of RL to end-to-end SGG by designing a graph-constrained reward to optimize structured outputs, while~\cite{li2025relation} incorporates a cognitive CoT mechanism to improve relation recognition. However, both approaches still fail to address the sparsity and long-tail issues of relations.

\subsection{Chain of Thought in Visual Reasoning}
The success of CoT in Large Language Models~\cite{wei2022chain} has motivated its adaptation to visual reasoning tasks. By breaking down complex problems into sequences of intermediate reasoning steps, CoT guides models to produce incremental solutions rather than direct answers. Several related methods~\cite{shao2024visual, huang2025lav,liu2025retrieval, chu2025qwen, yuan2026more} integrate explicit visual evidence during generation to bridge the gap between visual perception and logical reasoning. \cite{xu2025llava} adopt structured sequential stages to achieve marked improvements on reasoning-intensive tasks. Unlike general visual reasoning, SGG's dense prediction nature necessitates a tailored structured reasoning process.

\section{Methodology}
This section begins with a formal definition of the SGG task, followed by the introduction of our three-stage progressive reasoning process. Furthermore, we present a type-aware relation augmentation strategy designed for the SFT phase. Finally, we detail the stage-aligned reward functions and the RL algorithm integrated in our framework. The overall architecture is illustrated in Figure~\ref{fig:framework}.

\subsection{Problem Formulation}
The goal of SGG is to convert an input image $I$ into a structured graph representation $G = (\mathcal{V}, \mathcal{E})$. Here, $\mathcal{V}$ denotes the set of object nodes, where each node $v \in \mathcal{V}$ corresponds to an object instance represented by a tuple $(b, c)$. The vector $b \in \mathbb{R}^4$ describes the object's bounding box coordinates, and $c$ is its category label. The set $\mathcal{E}$ comprises relation triplets of the form $(v_s, p, v_o)$, where $v_s, v_o \in \mathcal{V}$ represent the subject and object of the triplet, respectively, and $p$ is the predicate describing their interaction. In a closed-set setting, all object categories $c$ and predicate categories $p$ are drawn from predefined sets $\mathcal{C}$ and $\mathcal{P}$, respectively.

\subsection{Three-stage Structured Reasoning}
The model acquires structured reasoning capabilities via visual instruction tuning in a prompt-response format. Guided by the theory of prompt space complexity~\cite{zhang2025prompt}, we design a CoT prompt that enforces explicit task constraints and a three-stage cognitive process (see Appendix~\ref{app:Three stage Structured Reasoning}). During SFT, the response is structured to systematize the unordered SGG process into a rigorous, sequential three-stage pipeline (details in Appendix~\ref{app:CoT Data Construction}). The structured reasoning stages are defined as follows.

\textbf{Stage 1: Object Category Detection.}
Direct bounding box detection for predefined object categories is prone to error propagation, resulting in imprecise localization, hallucination artifacts, and suppressed recall. To mitigate this, our model first identifies only the categories present in the image, effectively narrowing the search space to relevant object categories.

\textbf{Stage 2: Object Instance Grounding.}
Based on the category set from Stage 1, the model performs instance grounding by sequentially processing each identified category. Specifically, it detects and localizes all object instances belonging to the current category before proceeding to the next one.

\textbf{Stage 3: Multi-type Relation Extraction.}
This stage focuses on extracting \emph{(subject-predicate-object)} relation triplets. To capture the semantic and visual diversity of predicates, the model explores latent inter-object associations across three predefined predicate types. For example, in VG150~\cite{xu2017scene} we consider spatial, possessive, and interactive relations, while in PSG~\cite{yang2022panoptic} we distinguish spatial, static-interactive, and dynamic-interactive relations, with details in Appendix~\ref{app:predicate}. The output order of relations within each type adheres to the subject instance sequence in Stage 2.

To maintain structural integrity during reasoning, SGG-R\textsuperscript{3} explicitly marks each stage with dedicated tags: \texttt{<CATEGORY></CATEGORY>}, \texttt{<OBJECT></OBJECT>}, and \texttt{<RELATION></RELATION>}. These tags correspond to the start and end of each stage respectively, with outputs standardized in JSON format. All stages are completed in a single inference pass.

\subsection{Type-aware Relation Augmentation}
SFT enables the model to learn structured outputs and the target data distribution, yet its generalization is often constrained by the sparsity and long-tailed nature of relation annotations. To address this, we propose a type-aware relation augmentation strategy based on Qwen2.5-VL-32B~\cite{bai2025qwen2} to increase the number of relation instances for each predefined predicate type. 

Following the CoT prompt (see Appendix~\ref{app:Relation Augmentation}), the strategy first generates an image description conditioned on ground-truth entities. This description, alongside predefined predicate types, guides the subsequent generation of relation triplets. To ensure validity, rule-based heuristics enforce that subjects and objects align with ground-truth annotations and predicates strictly belong to the available category set. This approach substantially increases training examples and introduces semantic diversity (detailed analyses in Appendix~\ref{app:Dataset Analysis} and~\ref{app:Relation Augmentation Analysis}). To mitigate the generation of implausible relations, we maintain augmentation quality by encoding each generated triplet $t$ into an embedding $\mathbf{e}_t$ via Sentence-BERT~\cite{reimers2019sentence}. A triplet $t$ is retained exclusively when its cosine similarity with any ground-truth triplet embedding exceeds a threshold $\theta \in [0, 1]$:
\begin{equation}
\small
\text{A}(t) = \mathbb{I}\Big[ \exists \, t_{\mathrm{gt}} \in \mathcal{T}_{\mathrm{gt}}: \; \mathrm{sim}(\mathbf{e}_t, \mathbf{e}_{t_{\mathrm{gt}}}) \geq \theta \Big]
\end{equation}
where $\mathcal{T}_{\text{gt}}$ denotes the set of ground-truth triplets. This filtering mechanism improves the quality of SFT data, facilitating a generalizable model that provides robust initialization for subsequent RL.

\subsection{Reward Modeling}
Reward modeling is central to reinforcement learning with verifiable rewards (RLVR). We introduce a composite reward scheme meticulously aligned with the CoT pipeline, incorporating both process rewards to guide reasoning at each stage and format rewards to standardize structured output. This joint optimization simultaneously enhances visual grounding and relation extraction, facilitating robust end-to-end SGG.

\subsubsection{Format Reward}
The format reward $R_{\mathrm{format}}$ optimizes structural integrity by verifying the completeness of required tags and correctness of JSON syntax at each stage.

\subsubsection{Category Detection Reward}

The category recognition reward $R_{\mathrm{category}}$ is measured on the $\mathrm{F1}$ score between predicted and ground-truth category sets, encouraging comprehensive detection while penalizing over-prediction to mitigate reward hacking. Here, $P = \mathrm{TP} / (\mathrm{TP} + \mathrm{FP})$ and $R = \mathrm{TP} / (\mathrm{TP} + \mathrm{FN})$, where $\mathrm{TP}$, $\mathrm{FP}$, and $\mathrm{FN}$ denote true positives, false positives, and false negatives, respectively.  
\begin{equation}
\small
R_{\mathrm{category}} = \frac{2PR}{P+R}
\end{equation}

\subsubsection{Instance Grounding Reward}

The instance grounding reward $R_{\mathrm{node}}$ including $R_{\mathrm{box}}$ and $R_{\mathrm{recall}}$ is computed via bipartite matching between predicted and ground-truth object pairs, as detailed in the Appendix~\ref{app:experiment_config:dataset}. Following prior work~\cite{chen2025compile}, $R_{\mathrm{box}}$ combines $\mathrm{IoU}$ and $\mathrm{L_1}$ distance to measure bounding box accuracy. Recognizing the limitations of MLLMs in object detection recall~\cite{li2025lmm}, we introduce a recall-oriented reward $R_{\mathrm{recall}}$ to explicitly incentivize high-recall matching. For a ground-truth object $v_{j} = (b_{j}, c_{j})$ matched to a predicted object $v_{i} = (b_{i}, c_{i})$, the per-pair recall $r_j$ is formulated as:
\begin{equation}
\small
r_j = 
\begin{cases} 
1.0, & \mathrm{IoU}(b_{j}, b_{i}) > 0.5 \land (c_{j} = c_{i}) \\
0.5, & \mathrm{IoU}(b_{j}, b_{i}) > 0.5 \oplus (c_{j} = c_{i}) \\
0.0, & \mathrm{otherwise}
\end{cases}
\end{equation}
The overall instance grounding reward is then defined as the average of these per-pair scores across all ground-truth instances.

\begin{figure}[t]
  \includegraphics[width=1.0\linewidth]{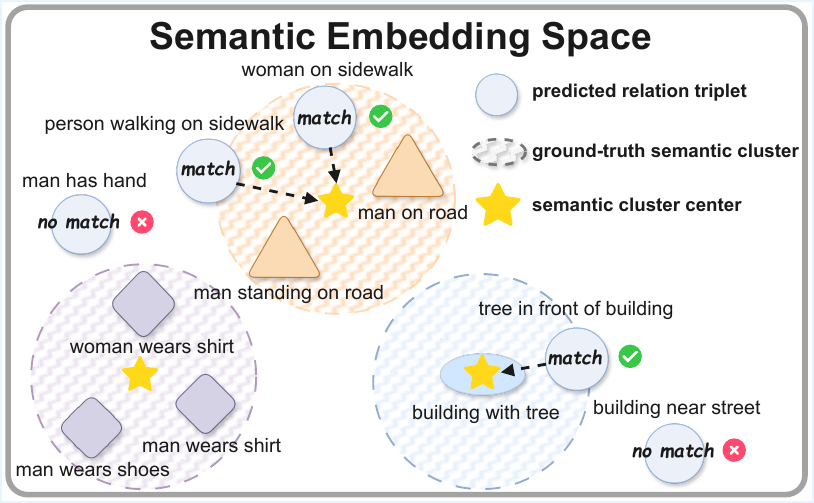} 
  \caption{Coarse-grained semantic clustering reward. A triplet is matched if its semantic embedding aligns with any ground-truth cluster centroid beyond a threshold, relaxing the strict matching requirement.}
  \label{fig:cluster}
\end{figure}

\subsubsection{Dual-granularity Reward}

The dual-granularity reward combines fine-grained and coarse-grained rewards to jointly optimize relation recall and semantic coverage, leveraging Sentence-BERT to obtain the relation embeddings. The fine-grained reward incorporates frequency-based adaptive weighting of predicates to alleviate long-tail bias in relation extraction. For each ground-truth triplet $t_j$ and its matched prediction $t_i$, let $\mathbf{e}_t$ and $\mathbf{e}_p$ denote the embeddings of the triplet and its predicate, respectively. The similarity $\mathrm{sim}_{\mathrm{j}}$ is calculated as:
\begin{equation}
\small
\mathrm{sim}_{\mathrm{j}} = \mathrm{sim}_{\mathrm{t}}(\mathbf{e}_{t_j}, \mathbf{e}_{t_i}) \times \mathrm{sim}_{\mathrm{p}}(\mathbf{e}_{p_j}, \mathbf{e}_{p_i})
\end{equation}
where $\mathrm{sim}_{\mathrm{t}}$ and $\mathrm{sim}_{\mathrm{p}}$ denote the cosine similarities of triplet and predicate embeddings, respectively (with $\mathrm{sim}_{\mathrm{j}} = 0$ for unmatched triplets). The fine-grained reward is formulated as:
\begin{equation}
\small
R_{\mathrm{fine}} = \frac{\sum_{t_j \in \mathcal{T}_{\mathrm{gt}}} \mathrm{sim}_{\mathrm{j}} \times w(p_j)}{\sum_{t_j \in \mathcal{T}_{\mathrm{gt}}} w(p_j)}
\end{equation}
where $w(p)$ is an adaptive weight based on the frequency of ground-truth predicate, defined as:
\begin{equation}
\small
    w(p) = w_{\mathrm{base}} + w_{\mathrm{inc}} \cdot \alpha(p)
\end{equation}
\begin{equation}
\small
    \alpha(p) = \frac{\log(1/f_p) - \log(1/f_{\mathrm{max}})}{\log(1/f_{\mathrm{min}}) - \log(1/f_{\mathrm{max}})}
\end{equation}
Here, $f_p$, $f_{\mathrm{max}}$, and $f_{\mathrm{min}}$ denote the frequency of predicate $p$, the maximum and the minimum predicate frequency in the training data. Meanwhile, $w_{\mathrm{base}}$ and $w_{\mathrm{inc}}$ represent the base weight and the incremental weight.

The coarse-grained reward $R_{\mathrm{coarse}}$ is designed to enhance model generalization beyond the limited sparse annotated data. It employs a cluster algorithm to encourage the model to generate relation triplets that are semantically proximate to ground-truth, without requiring exact subject and object matching, as illustrated in Figure~\ref{fig:cluster}. First, ground-truth triplets are embedded and clustered into semantic prototypes $\mathcal{C}_{\mathrm{GT}}$ using DBSCAN. Each resulting cluster $c_j \in \mathcal{C}_{\mathrm{GT}}$ is represented by the mean embedding $\mathbf{e}_{c_j}$ of its member triplets. The reward is decomposed into two complementary metrics. Cluster coverage $N_{\mathrm{covered}}$ measures the number of ground-truth clusters that are matched by at least one predicted triplet $t_i$, where the similarity threshold is set as $\tau \in (0, 1)$:
\begin{equation}
\small
N_{\mathrm{covered}} = \sum_{c_j \in \mathcal{C}_{\mathrm{GT}}} \mathbb{I}\left[ \exists \, t_i \in \mathcal{T}_{\mathrm{pred}}: \mathrm{sim}\big(\mathbf{e}_{t_i}, \mathbf{e}_{c_j}\big) \geq \tau \right]
\end{equation}
where $\mathcal{T}_{\mathrm{pred}}$ is the set of predicted triplets. Cluster density $\rho_{\mathrm{covered}}$ assesses the concentration of predictions within the matched clusters:
\begin{equation}
\small
\rho_{\mathrm{covered}} = \frac{\sum_{c_j \in \mathcal{C}_{\mathrm{covered}}} N_{\mathrm{pred}}^{c_j}}{\sum_{c_j \in \mathcal{C}_{\mathrm{covered}}} N_{\mathrm{gt}}^{c_j}}
\end{equation}
Here, $N_{\mathrm{pred}}^{c_j}$ and $N_{\mathrm{gt}}^{c_j}$ denote the number of predicted and ground-truth triplets in cluster $c_j$, respectively, and $\mathcal{C}_{\mathrm{covered}}$ is the subset of ground-truth clusters that are successfully matched. Finally, the coarse-grained reward combines coverage and density to form a unified metric:
\begin{equation}
\small
R_{\mathrm{coarse}} = \left( \frac{N_{\mathrm{covered}}}{|\mathcal{C}_{\mathrm{GT}}|} \right) \times \rho_{\mathrm{covered}}
\end{equation}

\subsection{Reinforcement Learning with Group Sequence Policy Optimization}

We adopt Group Sequence Policy Optimization (GSPO)~\cite{zheng2025group} as the online algorithm for the RL phase. To address the high-variance training noise and instability associated with token-level importance sampling in large-scale RL, the GSPO objective $\mathcal{J}_{\text{GSPO}}$ extends GRPO~\cite{shao2024deepseekmath} by adopting a sequence-level importance ratio $s_{i}(\theta)$, replacing the original token-level version, which is formulated as: 
\begin{equation}
\small
\label{eq:gspo-obj}
\begin{aligned}
&\mathcal{J}_{\text{GSPO}}(\theta) = \mathbb{E}_{x \sim \mathcal{D},\, \{y_{i}\}_{i=1}^{G} \sim \pi_{\theta_{\text{old}}}(\cdot|x)} \bigg[ \frac{1}{G} \sum_{i=1}^{G} \min\\
&  \Big( s_{i}(\theta) \widehat{A}_{i},\ \mathrm{clip}\big(s_{i}(\theta), 1-\varepsilon, 1+\varepsilon\big) \widehat{A}_{i} \Big) \bigg]
\end{aligned}
\end{equation}

Here, the sequence-level importance ratio $s_{i}(\theta)$ is defined as the geometric mean of the policy ratio $\pi_{\theta}/\pi_{\theta_{\text{old}}}$ over the entire output sequence $y_i$ for numerical stability. This sequence-level formulation enhances optimization robustness for the generation of long-form, structured JSON outputs.
\begin{equation} 
\small
\label{eq:si-ratio} 
s_{i}(\theta) = \exp \left( \frac{1}{|y_{i}|} \sum_{t=1}^{|y_{i}|} \log \frac{\pi_{\theta}(y_{i,t}|x, y_{i,<t})}{\pi_{\theta_{\text{old}}}(y_{i,t}|x, y_{i,<t})} \right) 
\end{equation}

The group-normalized advantage $\widehat{A}_i$ is retained as the advantage, and estimated by normalizing the reward $r(x, y_i)$ using the mean and standard deviation of rewards within the current group, without relying on a learned value function $V$:
\begin{equation}
\small
\label{eq:advantage}
\widehat{A}_{i} = \frac{r(x, y_{i}) - \mathrm{mean} \left( \{ r(x, y_{i}) \}_{i=1}^{G} \right) }{\mathrm{std} \left( \{ r(x, y_{i}) \}_{i=1}^{G} \right) }
\end{equation}

\section{Experiments}

\subsection{Experiment Setup}

\textbf{Datasets.} We evaluate our method on two standard SGG benchmarks. The first is the widely adopted Visual Genome (VG150) dataset~\cite{xu2017scene}, which includes 150 object categories and 50 relation categories, and follows the prior work~\cite{chen2024expanding, chen2025compile} to split image-graph pairs with 56,224 training, 5,000 validation, and 14,700 testing. The second is the Panoptic Scene Graph (PSG)  dataset~\cite{yang2022panoptic}, constructed based on COCO~\cite{lin2014microsoft}, containing 80 thing categories, 53 stuff categories, and 56 relation categories. It provides 46,563 training pairs and 2,186 test pairs for evaluation.

\noindent \textbf{Evaluation Metrics.} We follow the standard SGDET~\cite{xu2017scene} evaluation protocol for SGG, in which models must generate scene graphs directly from images without ground-truth object bounding boxes. The main evaluation metrics are Recall, mean Recall (mRecall), and zero-shot Recall (zsRecall). A triplet is correct only if both the subject and object bounding boxes have $\mathrm{IoU} \geq 0.5$ with their matched ground-truth boxes, and the subject, object, and predicate categories are all correctly predicted. Recall is computed among all predicate categories, mRecall averages the recall per predicate category, and zsRecall specifically measures the model’s generalization to relation triplets that are unseen during training. We adopt $K=100$ for all recall calculations.

\noindent \textbf{Implementation Details.} Our implementation is built upon the TRL~\cite{vonwerra2022trl} library, with vLLM~\cite{kwon2023efficient} for accelerated sampling during the RL stage. The training is conducted on 8 NVIDIA A800 (40GB) GPUs. Implementation details are provided in the Appendix~\ref{app:Experiment Configurations}.

\begin{table}[t]
  \scriptsize
  \renewcommand{\arraystretch}{0.9}
  \centering
   \begin{tabular}{l 
c | 
S[table-format=2.1,table-column-width=0.55cm] 
S[table-format=2.1,table-column-width=0.55cm] 
S[table-format=2.1,table-column-width=0.55cm] 
S[table-format=1.1,table-column-width=0.55cm]}
    \toprule
    \multirow{2}{*}{\textbf{Method}} & \multirow{2}{*}{\textbf{Params}} & \multicolumn{4}{c}{\textbf{VG150 Dataset}} \\
    \cmidrule(lr){3-6}
    & & \textbf{Recall} & \textbf{mRecall} & \textbf{Mean} & \textbf{zsRecall} \\
    \midrule
    \rowcolor{gray!15}
    \multicolumn{6}{l}{\textit{Non-VLM Methods}} \\
    MOTIFS  & {--} & 30.3 & 6.6 & 18.5 & 0.2 \\
    VCTree  & {--} & 31.3 & 8.0 & 19.7 & 0.7 \\
    GPS-Net & {--} & 31.7 & 9.8 & 20.8 & {--} \\
    PE-Net  & {--} & 35.2 & 14.5 & 24.9 & 3.6 \\
    RelTR  & {--} & 30.7 & 12.6 & 21.7 & 2.4 \\
    SGTR  & {--} & 28.4 & 15.2 & 21.8 & 5.8 \\
    SGTR+  & {--} & 30.1 & 17.0 & 23.6 & 5.8 \\
    HiLo  & {--} & 29.2 & \bfseries 17.6 & 23.4 & {--} \\
    Pair-Net  & {--} & 29.3 & 15.4 & 22.4 & {--} \\
    DSGG  & {--} & 38.5 & \underline{17.3} & \bfseries 27.9 & 3.9 \\
    EGTR & {--} & 34.3 & 10.1 & 22.2 & {--} \\
    \midrule
    \rowcolor{gray!15}
    \multicolumn{6}{l}{\textit{VLM-based Methods}} \\
    VS\textsuperscript{3}   & {--} & \underline{40.9} & 7.8 & 24.4 & {--} \\
    OvSGTR  & {--} & \bfseries 41.3 & 8.8 & 25.1 & {--} \\
    PGSG & {--}  & 23.6 & 12.7 & 18.2 & \bfseries 7.6 \\
    OwSGG$^\dag$ & 72B & 3.2 & 3.4 & 3.3 & 2.0 \\
    R1-SGG$^\dag$   & 7B & 27.6 & 10.9 & 19.3 & 4.4 \\
    \midrule
    \rowcolor{gray!15}
    \multicolumn{6}{l}{\textit{Ours}} \\
    SGG-R\textsuperscript{3}$^\ddagger$ (SFT)   & 3B & 17.5 & 8.2 & 12.9 & 2.4 \\
    SGG-R\textsuperscript{3}$^\ddagger$ (SFT + RL)  & 3B & 36.0 & 14.8 & \underline{25.4} & \underline{6.1} \\
    \bottomrule
  \end{tabular}
  \caption{Performance ($\%$) comparison of our proposed method against non-VLM and VLM-based baselines on the VG150 test set. Bold text indicates best performance, underlined text indicates second best. The ``Mean'' denotes the average of Recall and mean Recall. $^\dag$ and $^\ddagger$ denote Qwen2-VL and Qwen2.5-VL as the base model respectively.}
  \label{tab:vg}
\end{table}

\begin{table}[t]
  \scriptsize
  \renewcommand{\arraystretch}{0.9}
  \centering
  \begin{tabular}{l 
c | 
S[table-format=2.1,table-column-width=0.55cm] 
S[table-format=2.1,table-column-width=0.55cm] 
S[table-format=2.1,table-column-width=0.55cm] 
S[table-format=1.1,table-column-width=0.55cm]}
    \toprule
    \multirow{2}{*}{\textbf{Method}} & \multirow{2}{*}{\textbf{Params}} & \multicolumn{4}{c}{\textbf{PSG Dataset}} \\
    \cmidrule(lr){3-6}
    & & \textbf{Recall} & \textbf{mRecall} & \textbf{Mean} & \textbf{zsRecall} \\
    \midrule
    \rowcolor{gray!15}
    \multicolumn{6}{l}{\textit{Non-VLM Methods}} \\
    MOTIFS  & {--} & 22.0 & 9.1 & 15.6 & {--} \\
    VCTree  & {--} & 20.6 & 9.7 & 15.2 & {--} \\
    GPS-Net & {--} & 20.6 & 7.2 & 13.9 & {--} \\
    PSGFormer  & {--} & 21.0 & 19.8 & 20.4 & 4.9 \\
    PSGTR   & {--} & 35.3 & 21.5 & 28.4 & 6.4 \\
    SGTR  & {--} & 32.0 & 23.5 & 27.8 & 3.6 \\
    HiLo  & {--} & \underline{51.4} & 40.9 & 46.2 & {--} \\
    Pair-Net  & {--} & 42.4 & 29.7 & 36.1 & {--} \\
    DSGG  & {--} & 50.0 & \underline{43.4} & \underline{46.7} & {--} \\
    \midrule
    \rowcolor{gray!15}
    \multicolumn{6}{l}{\textit{VLM-based Methods}} \\
    OvSGTR  & {--} & 41.4 & 28.3 & 34.9 & {--} \\
    PGSG & {--}  & 33.4 & 22.1 & 27.8 & \bfseries 8.9 \\
    ASMv2  & 13B  & 14.8 & 11.8 & 13.3 &  {--} \\
    LLaVA-SpaceSGG  & 13B & 15.4 & 13.2 & 14.3 & {--} \\
    OwSGG$^\dag$ & 72B & 14.0 & 13.7 & 13.9 & 6.4 \\
    R1-SGG$^\dag$   & 7B & 43.5 & 33.2 & 38.4 & \underline{7.7} \\
    Relation-R1$^\ddagger$   & 3B & 25.9 & 21.3 & 23.6 & {--} \\
    \midrule
    \rowcolor{gray!15}
    \multicolumn{6}{l}{\textit{Ours}} \\
    SGG-R\textsuperscript{3}$^\ddagger$ (SFT)   & 3B & 33.3 & 26.1 & 29.7 & 0.0 \\
    SGG-R\textsuperscript{3}$^\ddagger$ (SFT + RL)  & 3B & \bfseries 52.5 & \bfseries 44.3 & \bfseries 48.4 & \underline{7.7} \\
    \bottomrule
  \end{tabular}
  \caption{Performance ($\%$) comparison of our proposed method against non-VLM and VLM-based baselines on the PSG test set. Bold text indicates best performance, underlined text indicates second best. The ``Mean'' denotes the average of Recall and mean Recall. $^\dag$ and $^\ddagger$ denote Qwen2-VL and Qwen2.5-VL as the base model respectively.}
  \label{tab:psg}
\end{table}

\subsection{Comparisons with State-of-the-Arts}

We benchmark our framework on the VG150 and PSG test sets against a comprehensive suite of baselines, including traditional non-VLM methods~\cite{zellers2018neural,tang2019learning,lin2020gps,yang2022panoptic, li2022sgtr,zheng2023prototype,cong2023reltr,zhou2023hilo,im2024egtr,li2024sgtr,wang2024pair,hayder2024dsgg} and VLM-based approaches~\cite{zhang2023learning,chen2024expanding,li2024pixels, wang2024all, xu2025llava,dutta2025open, chen2025compile, li2025relation}, as summarized in Table~\ref{tab:vg} and Table~\ref{tab:psg}.

We adopt Qwen2.5-VL-3B~\cite{bai2025qwen2} as the base model and conduct experiments under two settings: one trained solely with SFT, and another further optimized via RL. Initially, it can be observed that the model's performance after SFT is lower than that of previous methods on both datasets. However, after further RL training initialized with the SFT model, the model shows substantial improvements across all metrics, with qualitative comparison provided in Appendix~\ref{app:sft}. This indicates that RL significantly enhances the capability for relation extraction.

On VG150 shown in Table~\ref{tab:vg}, our method achieves the highest mRecall among all VLM-based approaches while maintaining a competitive Recall. Although our mRecall outperforms most baselines, it lags slightly behind some non-VLM methods~\cite{zhou2023hilo, wang2024pair, hayder2024dsgg}. We attribute this gap to the exhaustive object-pair coverage inherent in classification frameworks, which remains a persistent challenge for generative models. Notably, the mean of our Recall and mRecall ranks second among all evaluated methods. Our approach also surpasses all non-VLM models in zsRecall, demonstrating superior generalization and the ability to recognize diverse relations.

Furthermore, our method demonstrates leading performance on the PSG dataset, as detailed in Table~\ref{tab:psg}. It significantly outperforms all multimodal methods and marginally exceeds non-VLM approaches in terms of Recall and mRecall, indicating a substantial reduction in relation bias. The average of our Recall and mRecall metrics also ranks first among all evaluated methods, substantially surpassing most baselines. In terms of zsRecall, our model achieves highly competitive performance, although it falls slightly short of~\cite{li2024pixels}. In summary, our framework achieves leading overall performance against both non-VLM and VLM-based baselines on the two benchmarks.

\begin{table*}[t]
  \scriptsize
  \renewcommand{\arraystretch}{0.7}
  \centering
  \begin{tabular}{l l c | S[table-format=2.2] | S[table-format=2.2] S[table-format=2.2] S[table-format=2.2] | S[table-format=2.2] | S[table-format=2.2] S[table-format=2.2] S[table-format=2.2]}
    \toprule
    \multirow{2}{*}{\textbf{Method}} & \multirow{2}{*}{\textbf{Training}} & \multirow{2}{*}{\textbf{Params}} & \multicolumn{4}{c}{\textbf{VG150 Dataset}} & \multicolumn{4}{c}{\textbf{PSG Dataset}} \\
    \cmidrule(lr){4-7} \cmidrule(lr){8-11}
    & & & \textbf{Failure Rate} & \textbf{Recall} & \textbf{mRecall} & \textbf{Mean} & \textbf{Failure Rate} & \textbf{Recall} & \textbf{mRecall} & \textbf{Mean} \\
    \midrule
    \rowcolor{gray!15}
    \multicolumn{11}{l}{\textit{Baselines}} \\
    R1-SGG$^\dagger$  & SFT & 2B & 75.01 & 4.82 & 1.04  & 2.93    & 9.79 & 21.83 & 13.79 & 17.81 \\
    R1-SGG$^\ddagger$  & SFT & 3B & 63.16 & 6.60 & 1.61  & 4.11    & 6.30 & 22.63 & 14.60 & 18.62 \\
    R1-SGG$^\dagger$ & SFT & 7B & 39.54 & 9.62  & 3.30  & 6.46    & 0.96 & 24.73 & 17.11 & 20.92 \\
    R1-SGG$^\dagger$  & SFT + RL & 2B & 0.10  & 21.09 & 7.48  & 14.29   & 2.70 & 38.49 & 31.21 & 34.85 \\
    R1-SGG$^\dagger$ & SFT + RL & 7B & 0.12  & 23.75 & 11.43 & 17.59   & \bfseries 0.00 & 43.48 & 33.71 & 38.60 \\
    \midrule
    \rowcolor{gray!15}
    \multicolumn{11}{l}{\textit{Ours (w/o RA)}} \\
    SGG-R\textsuperscript{3}$^\dagger$ & SFT & 2B & 17.78 & 9.93  & 2.08  & 6.01    & 0.23 & 22.66 & 12.25 & 17.46 \\
    SGG-R\textsuperscript{3}$^\ddagger$ & SFT & 3B & 6.24  & 11.98 & 2.36  & 7.17    & 0.18 & 22.27 & 11.45 & 16.86 \\
    SGG-R\textsuperscript{3}$^\ddagger$ & SFT + RL & 3B & 0.12  & 27.85 & 12.68 & 20.27   & \bfseries 0.00 & 47.63 & 38.33 & 42.98 \\
    \midrule
    \rowcolor{gray!15}
    \multicolumn{11}{l}{\textit{Ours (w/ RA)}} \\
    SGG-R\textsuperscript{3}$^\dagger$ & SFT & 2B & 13.37 & 13.90 & 8.08  & 10.99   & 0.05 & 33.06 & 24.67 & 28.87 \\
    SGG-R\textsuperscript{3}$^\ddagger$ & SFT & 3B & 3.30  & 14.88 & 8.80  & 11.84   & 0.05 & 33.31 & 26.13 & 29.72 \\
    SGG-R\textsuperscript{3}$^\ddagger$ & SFT + RL & 3B & \bfseries 0.06  & \bfseries 30.66 & \bfseries 14.52  & \bfseries 22.59   & \bfseries 0.00 & \bfseries 52.49 & \bfseries 44.30 & \bfseries 48.40 \\
    \bottomrule
  \end{tabular}
  \caption{Performance ($\%$) comparison with MLLM-based end-to-end SGG baselines on the VG150 validation set and PSG test set. The ``Mean'' denotes the average of Recall and mean Recall. $^\dag$ and $^\ddagger$ denote Qwen2-VL and Qwen2.5-VL as the base model, respectively.}
  \label{tab:2}
\end{table*}

\begin{figure}[t]
  \includegraphics[width=1.0\linewidth]{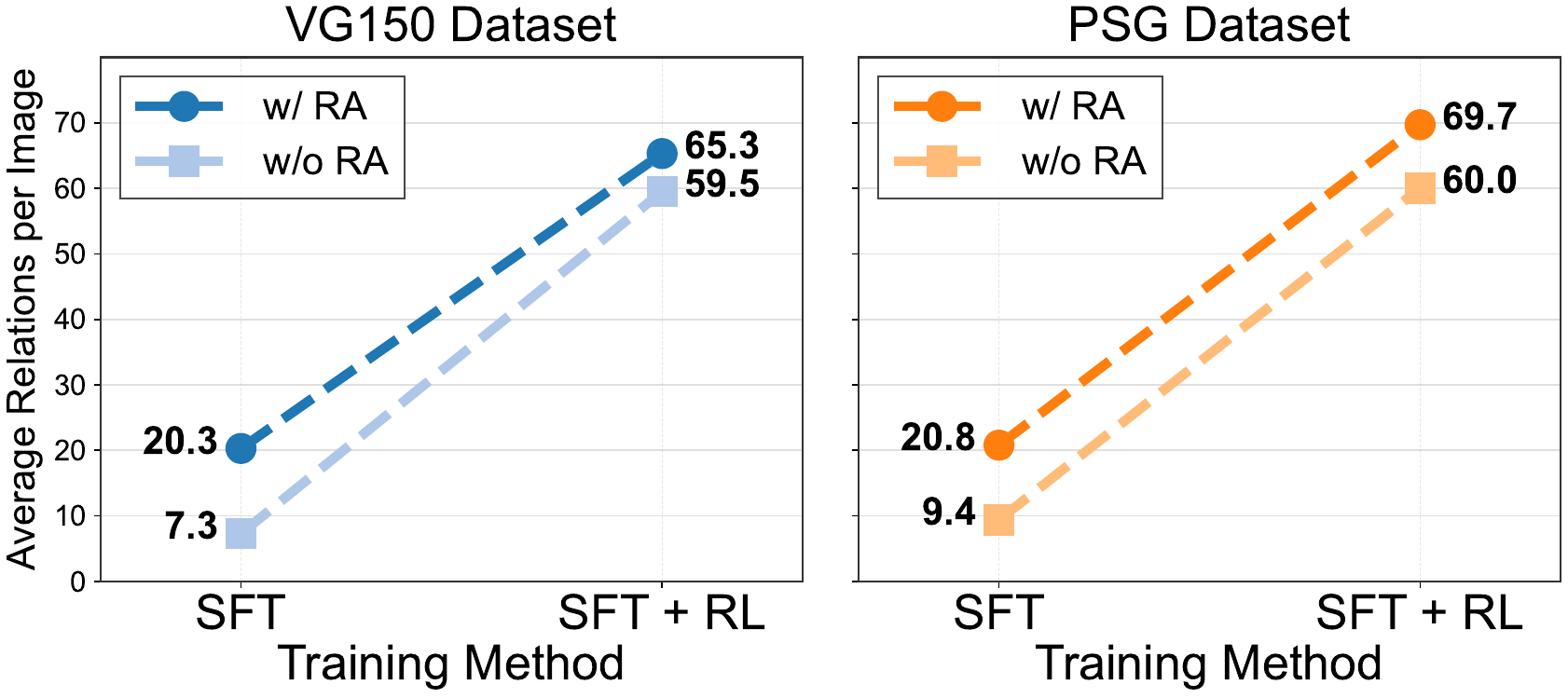}
  \caption{Quantitative analysis of the average number of relations generated per image on the VG150 and PSG test sets: SFT, SFT + RL; with/without relation augmentation (RA).}
  \label{fig:average number}
\end{figure}

\subsection{Comparisons with MLLM-based Methods} 

\textbf{Relation Recognition Evaluation.} As shown in Table~\ref{tab:2}, we compare our proposed method with state-of-the-art MLLM-based baselines~\cite{chen2025compile}, which directly generate objects and relations without a structured reasoning process. Following the baselines, we use the Failure Rate metric to assess the structural integrity and validity of the generated JSON format. For a fair comparison, we report results based on both Qwen2-VL~\cite{wang2024qwen2} and Qwen2.5-VL, evaluating the performance with and without our relation augmentation. Furthermore, a detailed analysis of MLLM zero-shot performance in SGG is provided in Appendix~\ref{app:MLLM Zero-Shot performance in SGG}.

When trained on equivalent data volumes, our 2B and 3B SFT models demonstrate superior robustness over baselines. Specifically, the 2B model achieves a $57.23\%$ reduction in VG150 failure rates, validating the CoT framework's efficacy in improving instruction following ability and structural correctness. While relation sparsity on PSG initially limits gains, subsequent RL optimization enables the 3B model to surpass the 7B baseline across all metrics, with mRecall increasing by $1.25\%$ on VG150 and $4.62\%$ on PSG. These results indicate that for models of comparable capacity, refined post-training is the primary performance driver.

During the SFT phase, relation augmentation further reduces the output failure rate. For instance, our 2B SFT model achieves a $26.17\%$ lower failure rate than the 7B baseline on VG150, suggesting a synergistic effect between the increased quantity of relational data and our structured reasoning process, which in turn enhances the structural integrity of the generated outputs. Specifically, our 2B and 3B SFT models demonstrate superior Recall and mRecall on both datasets. Subsequent RL training successfully maintains this performance advantage. Notably, our 3B model maintains a substantial lead over the 7B baseline: on VG150, it outperforms the baseline by $6.91\%$ in Recall and $3.09\%$ in mRecall, while on the PSG dataset, these gains are even more pronounced at $9.01\%$ and $10.59\%$, respectively. Qualitative visualizations of our SGG results are provided in Appendix~\ref{app:sggv}.

\noindent \textbf{Object Detection Evaluation.} In SGG, we adopt Recall and mean Recall as the primary metrics for evaluating object detection, rather than the standard Average Precision (AP). This choice stems from the task's requirement to identify all plausible objects (even unlabeled ones) to support relation prediction. Here, a predicted object is considered a match if its $\mathrm{IoU}$ with a ground-truth box $\geq 0.5$ and their categories match.

As reported in Table~\ref{tab:recall_comparison}, our SFT models significantly outperforms the baselines across all metrics on both datasets. For instance, the 3B model achieves gains of $19.14\%$ in Recall and $14.35\%$ in mRecall on the VG150 dataset, demonstrating superior object detection capabilities. Notably, RL optimization yields a substantial boost over the 3B SFT model, it simultaneously enhances both Recall and mRecall, surpassing the 7B baseline by $10.58\%$ in Recall and $9.31\%$ in mRecall on VG150. Meanwhile, the zero-shot performance of Qwen2.5-VL-32B falls short of both the baseline and our proposed approach. These results indicate that our CoT framework improves both the precision and coverage of object recognition. Furthermore, the category and instance grounding rewards during RL effectively guide the model toward identifying a more exhaustive set of valid object instances, thereby establishing a robust foundation for comprehensive scene graph generation.

\begin{table}[t]
  \tiny
  \renewcommand{\arraystretch}{0.7}
  \centering
  \begin{tabular}{@{}l@{\hspace{0.6em}}l@{\hspace{0.8em}}c@{\hspace{0.6em}} | S[table-format=2.2] S[table-format=2.2] | S[table-format=2.2] S[table-format=2.2]@{}}
    \toprule
    \multirow{2}{*}{\textbf{Method}} & \multirow{2}{*}{\textbf{Training}} & \multirow{2}{*}{\textbf{Params}} & \multicolumn{2}{c|}{\textbf{VG150}} & \multicolumn{2}{c}{\textbf{PSG}} \\
    \cmidrule{4-5} \cmidrule{6-7}
    & & & \textbf{Recall} & \textbf{mRecall} & \textbf{Recall} & \textbf{mRecall} \\
    \midrule
    \rowcolor{gray!15}
    \multicolumn{7}{@{}l}{\textit{Open-sourced MLLMs}} \\
    Qwen2.5-VL  & zero-shot & 32B & 24.84 & 25.39 & 35.66 & 39.52 \\
    \midrule
    \rowcolor{gray!15}
    \multicolumn{7}{@{}l}{\textit{Baselines}} \\
    R1-SGG$^\dagger$  & SFT & 2B & 10.72 & 14.76 & 47.43 & 46.72 \\
    R1-SGG$^\ddagger$  & SFT & 3B & 14.98 & 18.89 & 51.97 & 49.94 \\
    R1-SGG$^\dagger$  & SFT + RL & 7B & 39.49 & 38.26 & 52.08 & 52.94 \\
    \midrule
    \rowcolor{gray!15}
    \multicolumn{7}{@{}l}{\textit{Ours}} \\
    SGG-R\textsuperscript{3}$^\dagger$ & SFT & 2B  & 32.04 & 31.15 & 56.52 & 53.78 \\
    SGG-R\textsuperscript{3}$^\ddagger$ & SFT & 3B & 34.12 & 33.24 & 56.50 & 53.46 \\
    SGG-R\textsuperscript{3}$^\ddagger$ & SFT + RL & 3B & \bfseries 50.07 & \bfseries 47.57 & \bfseries 61.32 & \bfseries 59.27 \\
    \bottomrule
  \end{tabular}
  \caption{Object detection performance ($\%$) comparison of our method against the baselines on the VG150 and PSG test sets. $^\dag$ and $^\ddagger$ indicate the use of Qwen2-VL and Qwen2.5-VL as the base models.}
  \label{tab:recall_comparison}
\end{table}

\begin{table*}[t]
  \scriptsize
  \renewcommand{\arraystretch}{0.5} % 压缩行高
  \centering
  \begin{tabular}{@{}l l l | S[table-format=2.2, table-column-width=0.70cm] S[table-format=2.2, table-column-width=0.70cm] S[table-format=1.2, table-column-width=0.70cm] | S[table-format=2.2, table-column-width=0.60cm] S[table-format=2.2, table-column-width=0.60cm] S[table-format=1.2, table-column-width=0.60cm] | S[table-format=2.2, table-column-width=0.70cm] S[table-format=2.2, table-column-width=0.70cm] S[table-format=1.2, table-column-width=0.70cm] | S[table-format=2.2, table-column-width=0.60cm] S[table-format=2.2, table-column-width=0.60cm] S[table-format=2.2, table-column-width=0.60cm]@{}}
    \toprule
    \multicolumn{3}{c}{\textbf{Setup}} & \multicolumn{6}{c}{\textbf{VG150 Dataset}} & \multicolumn{6}{c}{\textbf{PSG Dataset}} \\
    \cmidrule(lr){1-3} \cmidrule(lr){4-9} \cmidrule(lr){10-15}
    \textbf{RA} & \textbf{DGR} & \textbf{Training} & \textbf{Recall} & \textbf{mRecall} & \textbf{zsRecall} & \textbf{Head} & \textbf{Body} & \textbf{Tail} & \textbf{Recall} & \textbf{mRecall} & \textbf{zsRecall} & \textbf{Head} & \textbf{Body} & \textbf{Tail} \\
    \midrule
    \rowcolor{gray!15}
    \multicolumn{15}{l}{\textit{Relation Augmentation (RA) Ablation}} \\
    \midrule
    \textcolor{gray}{\ding{53}} & \textcolor{gray}{\ding{53}} & SFT & 12.88 & 2.41 & 0.90 & 13.99 & 1.13 & 0.53 & 22.27 & 11.45 & 0.00 & 22.32 & 14.92 & 1.21 \\
    \textcolor{gray}{\ding{51}} & \textcolor{gray}{\ding{53}} & SFT & 17.74 & 8.23 & 2.35 & 18.67 & 9.90 & 4.96 & 33.31 & 26.13 & 0.00 & 32.24 & 33.04 & 16.01 \\
    \midrule
    \textcolor{gray}{\ding{53}} & \textcolor{gray}{\ding{51}} & SFT + RL & 31.57 & 12.62 & 5.03 & 32.95 & 14.90 & 4.76 & 47.63 & 38.33 & 0.00 & 46.98 & 43.10 & 24.77 \\
    \textcolor{gray}{\ding{51}} & \textcolor{gray}{\ding{51}} & SFT + RL & 35.95 & 14.78 & 6.11 & 36.98 & 15.58 & 8.13 & 52.49 & 44.30 & 7.69 & 50.95 & 50.84 & 34.14 \\
    \midrule
    \rowcolor{gray!15}
    \multicolumn{15}{l}{\textit{Dual-granularity Reward (DGR) Ablation}} \\
    \midrule
    \textcolor{gray}{\ding{51}} & \textcolor{gray}{\ding{53}} & SFT + RL & 34.55 & 13.94 & 5.01 & 35.58 & 15.07 & 7.22 & 50.40 & 42.26 & 7.21 & 49.25 & 46.61 & 32.33 \\
    \textcolor{gray}{\ding{51}} & \textcolor{gray}{\ding{51}} & SFT + RL & 35.95 & 14.78 & 6.11 & 36.98 & 15.58 & 8.13 & 52.49 & 44.30 & 7.69 & 50.95 & 50.84 & 34.14 \\
    \bottomrule
  \end{tabular}
  \caption{Performance ($\%$) comparison for ablation studies of relation augmentation (RA) and dual-granularity reward (DGR) modules. \textcolor{gray}{\ding{51}} indicates the component is enabled, \textcolor{gray}{\ding{53}} indicates the component is disabled.}
  \label{tab:ablation_studies}
\end{table*}

\subsection{Ablation Study}

To verify the efficacy of the proposed relation augmentation (RA) and dual-granularity reward (DGR), we conduct extensive ablation studies. Specifically, we categorize predicates into Head, Body, and Tail classes based on training frequency to evaluate each module's contribution to long-tail relation recognition. Details regarding these partitions are provided in Appendix~\ref{app:evaluation setup}.

\noindent \textbf{Effectiveness of Relation Augmentation.} Table~\ref{tab:ablation_studies} shows that RA significantly boosts performance in both SFT and RL stages. Specifically, RA enhances the SFT model's ability to recognize body and tail predicates, yielding gains of $8.77\%$ and $4.43\%$ on VG150, and $18.12\%$ and $14.80\%$ on PSG. Building on this RA-enhanced initialization, RL further elevates Recall and mRecall, with tail performance increasing by $3.37\%$ on VG150 and $9.37\%$ on PSG. These findings suggest that RA provides a superior foundation for RL to explore long-tail relations. Notably, while RL training without RA fails to achieve any zsRecall on PSG, its incorporation yields a $7.69\%$ improvement, which underscores RA's pivotal role in facilitating a generalized ``cold start''.

Additionally, Figure~\ref{fig:average number} compares the average volume of generated relations per image with and without RA. Experimental results demonstrate that while the inherent sparsity of the raw data limits relation density, RA significantly promotes relation generation. Specifically, RA-enhanced models output an additional 13.0 and 11.4 relations per image on VG150 and PSG after SFT. Furthermore, RL training further stimulates reasoning depth, leading to a continued increase in the total number of relations. Throughout the RL phase, the model initialized with RA consistently maintains a higher relation density than its non-RA counterpart across both datasets.

\noindent \textbf{Impact of Dual-granularity Reward.} We evaluate the impact of DGR by comparing it against a baseline utilizing a standard relation recall reward. As shown in Table~\ref{tab:ablation_studies}, with the RA-enhanced SFT model as initialization, DGR yields consistent gains across both Recall and mRecall. For instance, on the PSG dataset, DGR improves Recall by $2.09\%$ and mRecall by $2.04\%$, demonstrating its effectiveness in mitigating the long-tail distribution of relations. Specifically, DGR facilitates balanced improvements across all predicate types, thereby mitigating generative bias. Furthermore, by leveraging semantic clustering to guide the optimization process, DGR enhances the model's generalization to novel or rare relational contexts, as evidenced by a $1.10\%$ increase in zsRecall on the VG150 dataset.

\section{Conclusion}

This paper presents SGG-R\textsuperscript{3}, a structured reasoning framework that aligns next-token prediction with end-to-end unbiased scene graph generation via a two-phase training pipeline. We propose a structured reasoning process encompassing category detection, instance grounding, and multi-type relation extraction. To generate comprehensive and unbiased scene graphs, we employ a relation-augmentation strategy to mitigate the cold-start challenge during SFT and introduce a dual-granularity reward scheme during RL. Empirical evaluations on the VG150 and PSG benchmarks demonstrate that SGG-R\textsuperscript{3} achieves superior capabilities of visual grounding and relation extraction, exhibiting significant potential for diverse downstream applications.

\section*{Limitations}

The proposed framework has limitations in efficiency, data fidelity, and label coverage. First, the use of MLLMs leads to non-negligible inference latency, which may constrain its use in real-time applications. Second, although we use filtering strategies, the relation augmentation process cannot guarantee entirely accurate triples; some erroneous or hallucinated data may still persist. The quality of this augmented data essentially sets the upper bound for model performance. Finally, our training is currently limited to a closed-set vocabulary, which may restrict the model's ability to recognize novel relations in the wild. Future work will focus on extending this framework to open-vocabulary settings to improve its generalization across diverse, real-world scenarios.

\section*{Acknowledgements}
This work was supported in part by the National Key Research and Development Program of China (Grant No. 2023YFC3304902).


%\clearpage
%\newpage
\bibliography{main}

\clearpage
\newpage
\appendix
\section*{Appendix}

\section{Implementation Details}
\label{sec:appendix}

\subsection{Experiment Configurations}
\label{app:Experiment Configurations}
Our experiments are conducted on 8 NVIDIA A800 (40GB) GPUs. During the supervised fine-tuning (SFT) stage, we employ the AdamW optimizer with a cosine learning rate scheduler, setting the initial learning rate to $1\times10^{-5}$ for training 1 epoch. The per-device batch size is set to 2 with 2 gradient accumulation steps, resulting in an effective total batch size of 32 ($2 \times 2 \times 8$). 

For the reinforcement learning (RL) phase, we utilize six GPUs for training and two GPUs for vLLM deployment to accelerate inference, using a sampling size of 8. The total batch size for RL is 96 ($8 \times 2 \times 6$). The model is trained for 1 epoch using the AdamW optimizer with a initial learning rate of $5\times10^{-7}$; further hyperparameter details are provided in Table~\ref{tab:training_parameters}. To compute semantic embeddings, we employ the all-MiniLM-L6-v2~\cite{reimers2019sentence} model to project relation triplets or predicates into 384-dimensional embeddings.

The similarity threshold for filtering augmented relation triplets is set to 0.9. The base weight based on predicate frequency is set to 1, and the incremental weight is 2. When applying DBSCAN to cluster ground-truth triplets, we empirically set the clustering threshold $\varepsilon = 0.75$ and $\texttt{min\_samples} = 1$ based on the distribution characteristics of the semantic space. During the evaluation phase, the similarity matching threshold between predicted triplets and their corresponding ground-truth clusters is likewise set to 0.75.

\subsection{Evaluation Setup}
\label{app:evaluation setup}
In the evaluation phase, we categorize predicates into \textbf{Head}, \textbf{Body}, and \textbf{Tail} based on their frequency in the training dataset to precisely assess model performance under long-tail distributions. Specifically, the top $30\%$ most frequent predicates are designated as head, the middle $30\%$ as body, and the bottom $40\%$ as tail. We then evaluate the recall performance within each group. Detailed statistics regarding the predicate counts and frequencies for each category are provided in Table~\ref{tab:longtail_distribution}.

\begin{table}[t]
  \centering
  \footnotesize 
  \renewcommand{\arraystretch}{1.2} 
  \begin{tabular}{@{}l l@{}} 
    \toprule
    \textbf{Hyperparameter} & \textbf{Value} \\
    \midrule
    Precision & bf16 \\ 
    Max completion length & 2048 \\
    Per-device batch size & 8 \\
    Gradient accumulation steps & 2 \\
    Scheduler type & Cosine \\
    Temperature & 1.0 \\
    Top-$p$ & 0.9 \\
    Top-$k$ & 50 \\
    Group size  & 8 \\
    Num iterations per generation & 1 \\
    Beta ($\beta$) & 0.0 \\
    Random seed  & 42 \\
    \bottomrule
  \end{tabular}
  \caption{Hyperparameter settings for reinforcement learning training.}
  \label{tab:training_parameters}
\end{table}

\begin{table}[t]
  \footnotesize
  \renewcommand{\arraystretch}{0.5} % 压缩行高
  \centering
  \begin{tabular}{@{}l | S[table-format=2.0] S[table-format=2.5] | S[table-format=2.0] S[table-format=2.5]@{}}
    \toprule
    \multirow{2}{*}{\textbf{Category}} & \multicolumn{2}{c}{\textbf{VG150}} & \multicolumn{2}{c}{\textbf{PSG}} \\
    \cmidrule(lr){2-3} \cmidrule(lr){4-5} 
    & \textbf{Count} & \textbf{Freq.\%} & \textbf{Count} & \textbf{Freq.\%} \\
    \midrule
    Head  & 15 & 94.60 & 16 & 93.97 \\
    Body  & 15 & 3.95  & 16 & 5.32 \\
    Tail  & 20 & 1.45  & 24 & 0.71 \\
    \bottomrule
  \end{tabular}
  \caption{Distribution of long-tail predicate types across two datasets. The table shows the number and frequency of predicates in each category: Head (top $30\%$), Body (middle $30\%$), and Tail (bottom $40\%$).}
  \label{tab:longtail_distribution}
\end{table}

\subsection{Semantic Taxonomy of Predicates}
\label{app:predicate}

Recognizing that distinct relations necessitate specific semantic and visual cues and acknowledging that a single object pair may have multiple concurrent relationships depending on different perspectives, we partition the predicates of each dataset into three distinct taxonomies based on their underlying semantic properties. For VG150, relations are classified into \textbf{spatial} (e.g., \textit{on, near}), \textbf{possessive} (e.g., \textit{has, belong to}), and \textbf{interactive} (e.g., \textit{wearing, standing on}) types. Similarly, for the PSG dataset, we distinguish between \textbf{spatial} (e.g., \textit{on, in}), \textbf{static-interactive} (e.g., \textit{sitting on, parked on}), and \textbf{dynamic-interactive} (e.g., \textit{walking on, jumping over}) relations.

\subsection{Structured Reasoning Prompt}
\label{app:Three stage Structured Reasoning}

\begin{figure*}[h]
  \centering
  \includegraphics[width=0.8\linewidth]{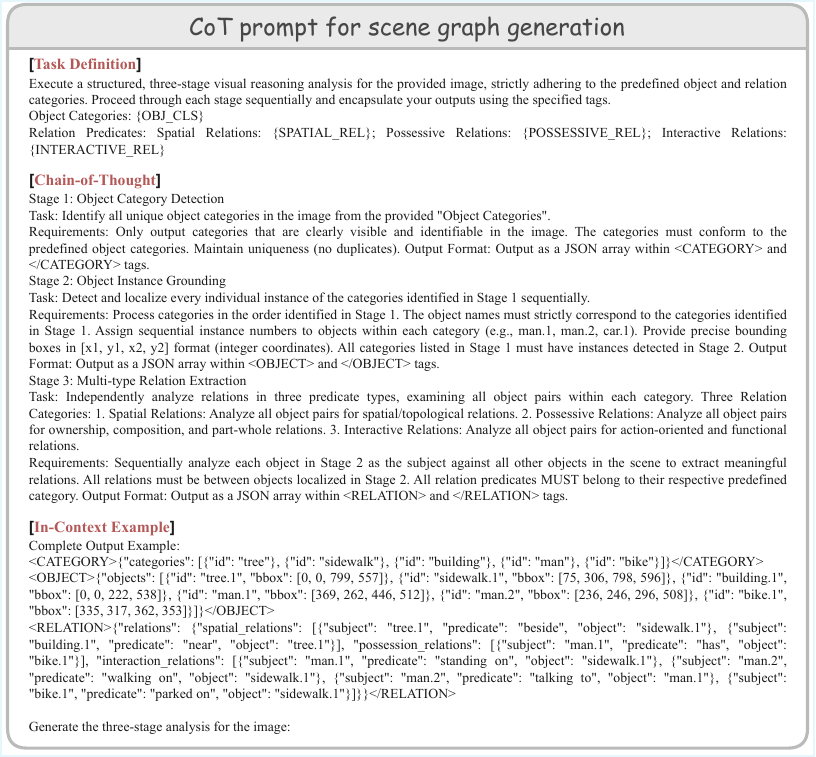}
  \caption{Three-stage CoT prompt template for end-to-end scene graph generation. The template is applied to both VG150 and PSG datasets, with the predefined object classes and predicate categories adjusted accordingly for each benchmark.}
  \label{fig:prompt1}
\end{figure*}

We design a three-stage Chain-of-Thought (CoT) prompt (see Figure~\ref{fig:prompt1}) to direct the structured output process of end-to-end Scene Graph Generation (SGG). In closed-set scenarios, the prompt incorporates the full set of object categories and relation predicate categories of three predefined types from the dataset. Specifically, the prompt guides the model through a sequential reasoning process based on these predefined categories, with each stage explicitly defining task objectives and constraints. To ensure robust parsing, all outputs are strictly constrained to standardized JSON formats within dedicated tags. Furthermore, we leverage in-context learning with representative examples to enhance the model's adherence to these structural constraints.

\begin{table*}[t]
  \tiny
  \centering
  \renewcommand{\arraystretch}{1.0}
  \setlength{\tabcolsep}{3.8pt} 
  
  \begin{tabular}{@{}l c | *{4}{S[table-format=6.0] S[table-format=5.0]} | *{4}{S[table-format=6.0] S[table-format=5.0]} @{}}
    \toprule
    \multirow{3}{*}{\textbf{Dataset}} & \multirow{3}{*}{\textbf{O/I}} & \multicolumn{8}{c|}{\textbf{Before RA}} & \multicolumn{8}{c}{\textbf{After RA}} \\
    \cmidrule(lr){3-10} \cmidrule(l){11-18}
    
    & & \multicolumn{2}{c}{\textbf{Total}} & \multicolumn{2}{c}{\textbf{Head}} & \multicolumn{2}{c}{\textbf{Body}} & \multicolumn{2}{c|}{\textbf{Tail}} & \multicolumn{2}{c}{\textbf{Total}} & \multicolumn{2}{c}{\textbf{Head}} & \multicolumn{2}{c}{\textbf{Body}} & \multicolumn{2}{c}{\textbf{Tail}} \\
    \cmidrule(lr){3-4} \cmidrule(lr){5-6} \cmidrule(lr){7-8} \cmidrule(lr){9-10} \cmidrule(lr){11-12} \cmidrule(lr){13-14} \cmidrule(lr){15-16} \cmidrule(l){17-18}
    
    & & {Count} & {Type} & {Count} & {Type} & {Count} & {Type} & {Count} & {Type} & {Count} & {Type} & {Count} & {Type} & {Count} & {Type} & {Count} & {Type} \\
    \midrule
    
    VG150 & 11.7 & 269196 & 25151 & 254656 & 20592 & 10645 & 2981 & 3985 & 1578 & 976465 & 30928 & 827105 & 24742 & 62374 & 3320 & 86986 & 2866 \\
    PSG   & 11.0 & 261666  & 18126  & 245886 & 16430 & 13932 & 1287 & 1848 & 409 & 810219 & 23530 & 742387 & 20542  & 55220  & 2150 & 12612 & 838 \\
    \bottomrule
  \end{tabular}
  
  \caption{Statistics of relation augmentation on VG150 and PSG training sets, with breakdowns for head/body/tail categories. ``O/I'' denotes average objects per image; ``Count'' refers to the number of relation triplets; ``Type'' denotes the number of relation triple combination patterns.}
  \label{tab:relation_sparsity}
\end{table*}

\subsection{Relation Augmentation Prompt}
\label{app:Relation Augmentation}

\begin{figure*}[h]
  \centering
  \includegraphics[width=0.8\linewidth]{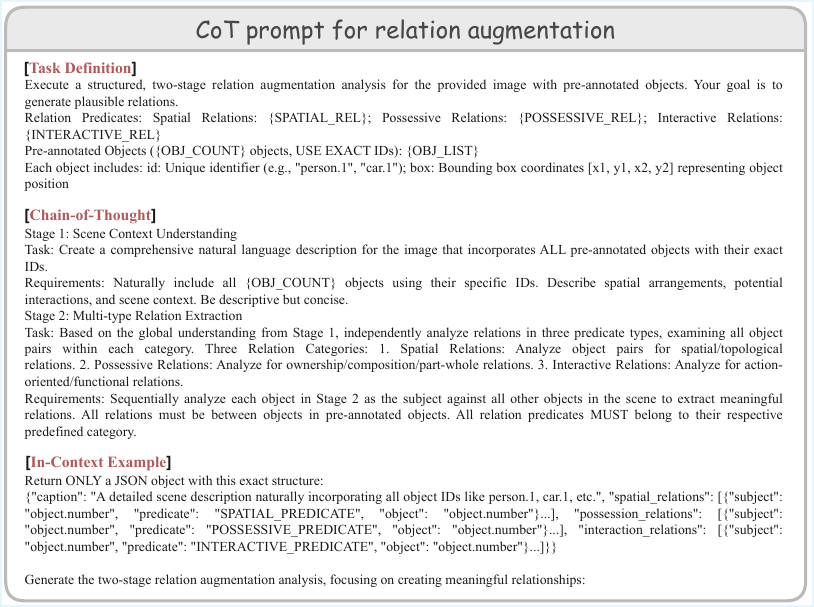}
  \caption{Two-stage CoT prompt template for relation augmentation. The template is applied to both VG150 and PSG datasets, with the predefined object list and predicate categories adjusted accordingly for each benchmark.}
  \label{fig:prompt2}
\end{figure*}

To mitigate the inherent sparsity of relation annotations, we leverage Qwen2.5-VL-32B for data augmentation. Our CoT prompting framework (see Figure~\ref{fig:prompt2}) integrates predicate categories of three predefined types and ground-truth object sets, including category names and bounding boxes. The prompt is structured into a sequential two-stage reasoning process, where each stage explicitly defines task objectives and requirements. To ensure precise instruction following and structural consistency, we also supplement representative in-context examples.

\subsection{CoT Data Construction}
\label{app:CoT Data Construction}
\begin{figure*}[t]
  \includegraphics[width=1.0\linewidth]{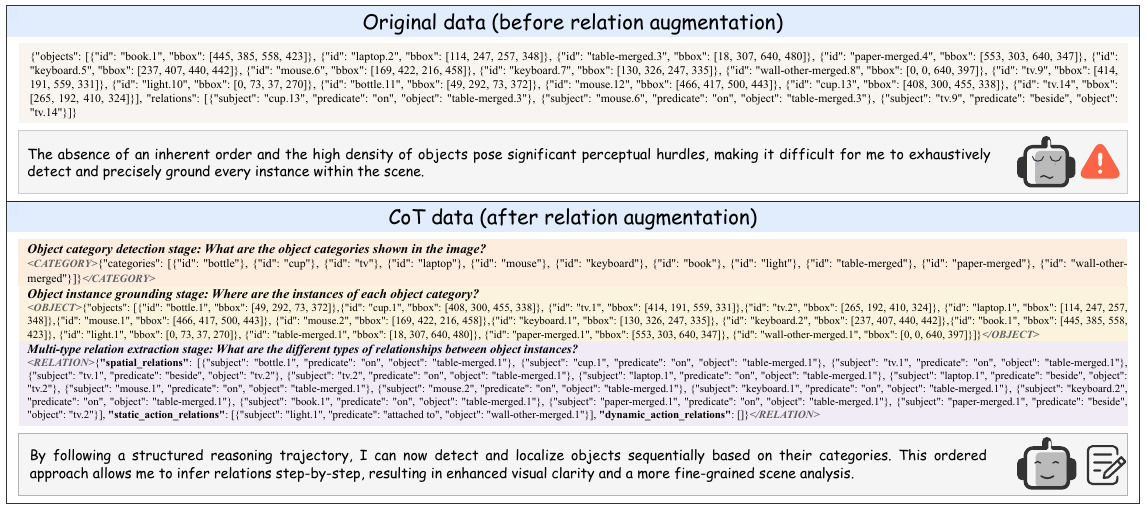}
  \caption{Comparison between original and CoT data. The original data are transformed into our three-stage format through a structured curation process.}
  \label{fig:cot}
\end{figure*}

We derive the CoT sequences from the original object annotations and augmented relation annotations. The data is organized into a three-stage sequential structure as illustrated in Figure~\ref{fig:cot}:

\textbf{Stage 1: Category Identification.} We extract unique object categories from the object instance set to construct an initial list. Given $M$ object instances $\mathcal{V} = \{(c_i, b_i)\}_{i=1}^M$, where $c_i$ and $b_i$ denote the category and bounding box respectively, we derive a unique category set $\mathcal{C} = \{c_n\}_{n=1}^N$ ($N \leq M$).

\textbf{Stage 2: Sequential Instance Grounding.} The instances in $\mathcal{V}$ are rearranged to align with the order of $\mathcal{C}$. To distinguish between multiple instances of the same category, we append unique numerical suffixes (e.g., \textit{person.1, person.2}) to each instance.

\textbf{Stage 3: Multi-relational Mapping.} Finally, the set of $K$ relation triplets $\mathcal{R} = \{(s_j, p_j, o_j)\}_{j=1}^K$ is organized. Triplets are first grouped by their predefined predicate types and then sorted based on the sequential order of subjects $s_j$ established in Stage 2. This alignment ensures that relational reasoning consistently follows the grounded object sequence.

To maintain clarity throughout the reasoning process, we explicitly encapsulates each stage within dedicated tags, denoting the start and end of each stage: \texttt{<CATEGORY></CATEGORY>}, \texttt{<OBJECT></OBJECT>}, and \texttt{<RELATION></RELATION>}. All outputs are standardized in JSON format within these tags. This sequential organization aligns with the instruction structure of the CoT prompt.

\subsection{Bipartite Matching}
\label{app:experiment_config:dataset}

Following~\cite{chen2025compile}, the assignment between predicted and ground-truth entities is formulated as a bipartite matching problem. The objective is to find an optimal mapping between the set of predicted nodes $\mathcal{V}_{pred} = \{v_i = (c_i, b_i)\}_{i=1}^M$ and ground-truth nodes $\mathcal{V}_{gt} = \{v_j = (c_j, b_j)\}_{j=1}^N$, where each node consists of an object class $c$ and its corresponding bounding box $b$. The matching cost between $v_i$ and $v_j$ is defined as:

\begin{equation}
\begin{split}
\mathcal{L}_{match}(v_i, v_j) = & \lambda_1 \left( 1 - \text{sim}(\mathbf{e}_{c_i}, \mathbf{e}_{c_j}) \right) \\
& + \lambda_2 \left( 1 - \text{IoU}(b_i, b_j) \right) \\
& + \lambda_3 \|b_i - b_j\|_1
\end{split}
\end{equation}

where $\mathrm{sim}(\mathbf{e}_{c_i}, \mathbf{e}_{c_j})$ denotes the cosine similarity between semantic embeddings of object classes extracted via Sentence-BERT, $\mathrm{IoU}(\cdot)$ is the Intersection over Union, and $\|\cdot\|_1$ represents the $\mathrm{L_1}$ distance. The hyper-parameters $\lambda_1, \lambda_2,$ and $\lambda_3$ serve as balancing coefficients that weight the relative contributions of semantic consistency, spatial overlap, and coordinate proximity, respectively.

To minimize false positives and enhance matching precision, we implement a threshold-based filtering mechanism. Only candidate pairs below a predefined cost threshold are retained for final assignment, which is determined by minimizing the global cost. This filtering provides a robust foundation for subsequent instance grounding optimization.

\section{Additional Analysis and Results}
\label{app:additional_analysis}

\subsection{Dataset Analysis}
\label{app:Dataset Analysis}

\noindent \textbf{Long-tail Analysis.} Table~\ref{tab:longtail_distribution} details the predicate distribution across both datasets, revealing a severe long-tail imbalance. In VG150, the head predicates comprising only $30\%$ of the predicate categories, but accounting for over $90\%$ of all training samples, leaving the body and tail categories significantly underrepresented. This extreme skewness highlights the substantial distribution bias inherent in these benchmarks, necessitating targeted strategies to improve rare category recognition.

\noindent \textbf{Sparsity Analysis.} As reported in Table~\ref{tab:relation_sparsity}, while the original datasets average approximately 11 objects per image, relation density remains disproportionately low at 4.8 for VG150 and 5.6 for PSG. This disparity indicates that most valid inter-object associations are unannotated, creating a sparse scene graph that hinders MLLM fine-tuning. Following our RA strategy, the average relation count rises to 17.4 per image, and the number of relations increases across all frequency groups. By generating plausible annotations for previously neglected pairs, our approach substantially enriches the relational supervision for the SFT phase.

\noindent \textbf{Diversity Analysis.} As shown in Table~\ref{tab:relation_sparsity}, the results reveal that RA not only expands the total volume of relations but also significantly enriches the variety of triplet types across all frequency groups. This demonstrates its effectiveness in enhancing semantic diversity and provides more generalized triplet combination patterns for model learning in the SFT stage.

\begin{figure}[t]
  \centering
  
  % 子图a (顶部)
  \begin{subfigure}[b]{1.0\linewidth}
    \centering
    \includegraphics[width=1.0\linewidth]{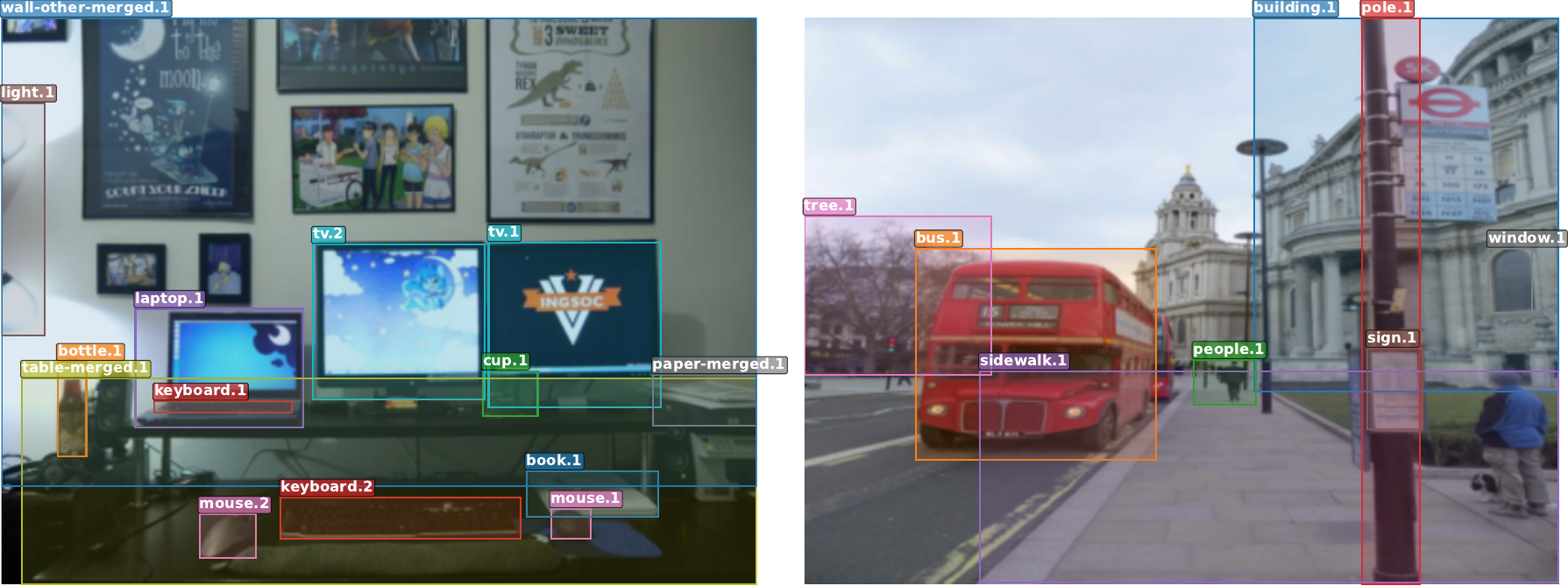}
    \caption{Ground-truth image samples with object bounding boxes for \textbf{PSG (left)} and \textbf{VG150 (right)}.}
    \label{fig:subfig_a}
  \end{subfigure}
  \\[0.8em]
  \begin{subfigure}[b]{1.0\linewidth}
    \centering
    \includegraphics[width=1.0\linewidth]{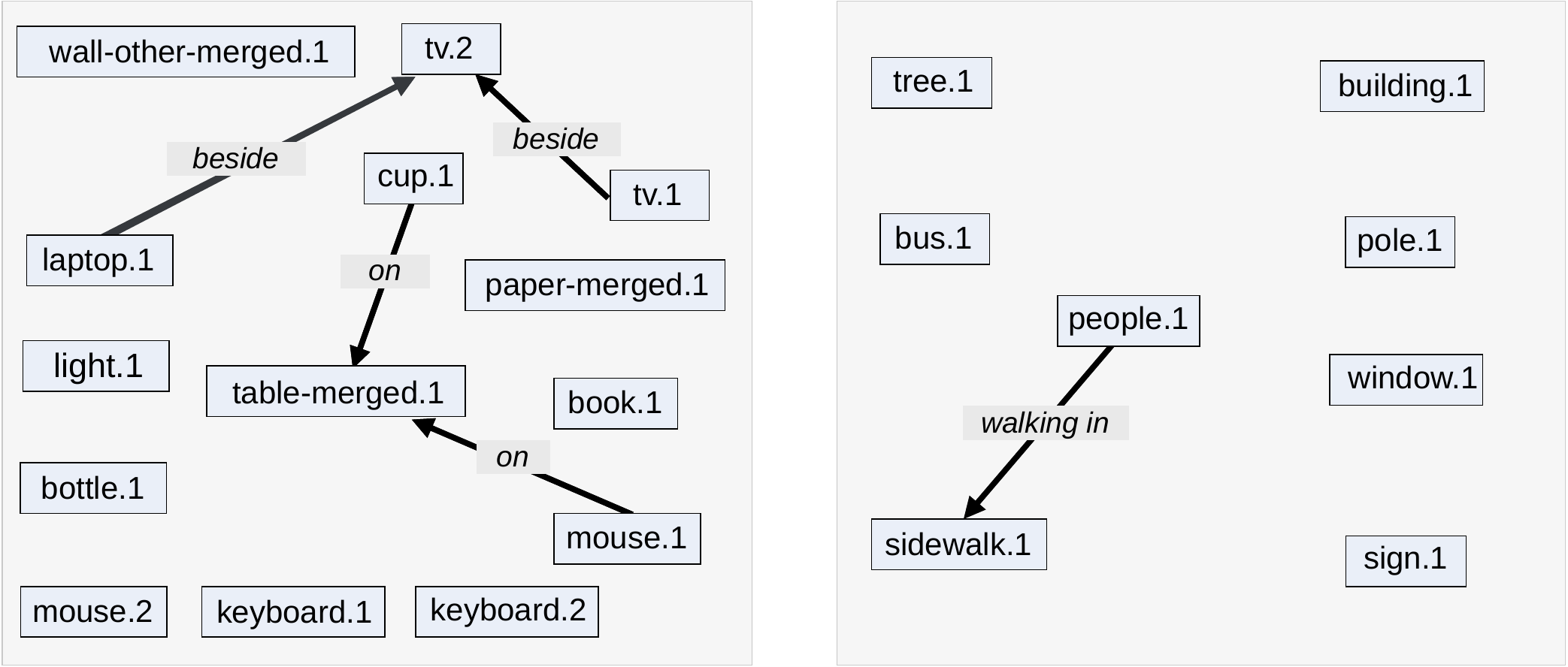}
    \caption{Original sparse scene graphs, where rectangles denote objects and directed edges represent relations.}
    \label{fig:subfig_b}
  \end{subfigure}
  \\[0.8em]
  \begin{subfigure}[b]{1.0\linewidth}
    \centering
    \includegraphics[width=1.0\linewidth]{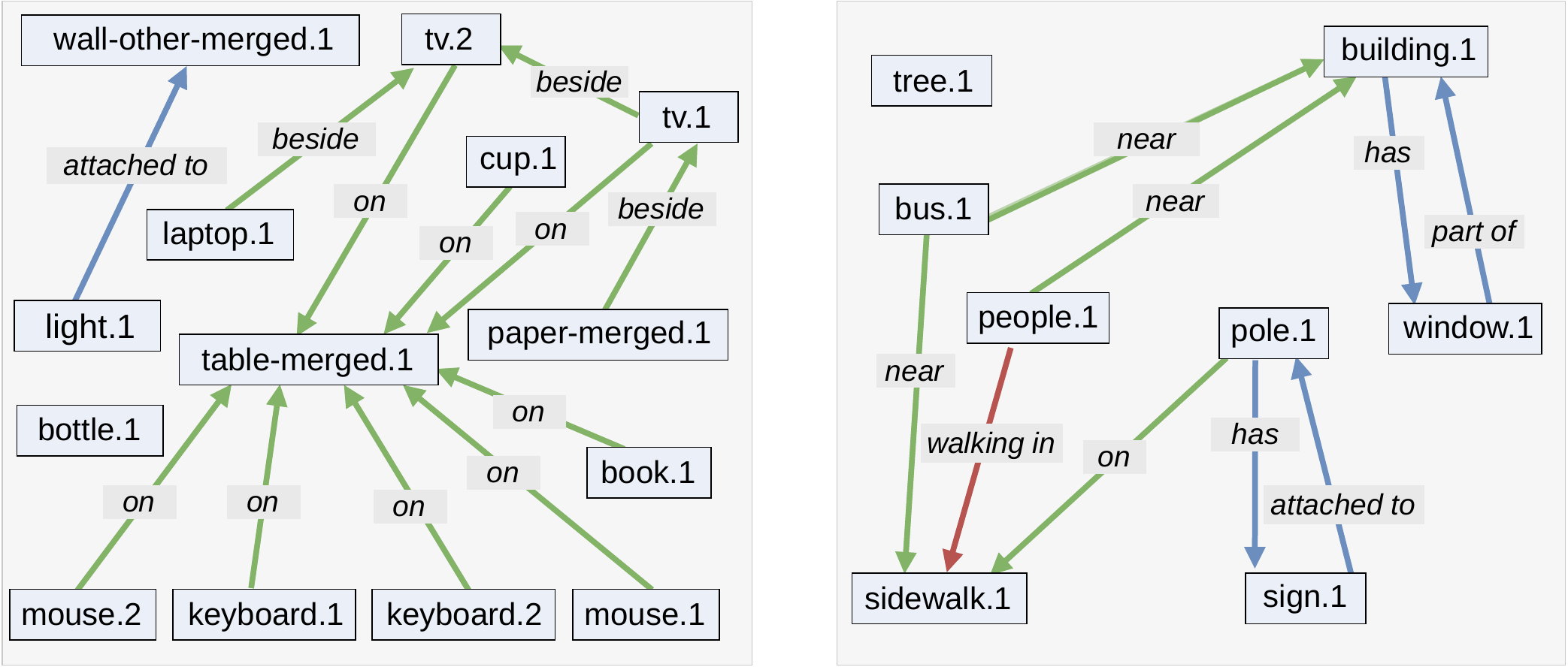}
    \caption{Augmented scene graphs with color-coded edges: for PSG, \textcolor{green}{green} and \textcolor{blue}{blue} edges signify \textit{spatial} and \textit{static-interactive} relations; for VG150, \textcolor{green}{green}, \textcolor{blue}{blue}, and \textcolor{red}{red} edges represent \textit{spatial}, \textit{possessive}, and \textit{interactive} relations, respectively.}
    \label{fig:subfig_c}
  \end{subfigure}
  
  \caption{Visualization of relation augmentation. (a) Original images with ground-truth object bounding boxes and category labels. (b) Visualization of original sparse relation annotations. (c) Visualization of the augmented relation annotations.}
  \label{fig:ra}
\end{figure}

\begin{table*}[t]
  \tiny
  \renewcommand{\arraystretch}{0.8} % 压缩行高
  \centering
\begin{tabular}{@{}l c|S[table-format=2.2, table-column-width=0.95cm]|S[table-format=2.2, table-column-width=0.58cm]S[table-format=2.2, table-column-width=0.58cm]S[table-format=1.2, table-column-width=0.58cm]|S[table-format=2.2, table-column-width=0.42cm]S[table-format=2.2, table-column-width=0.42cm]S[table-format=1.2, table-column-width=0.42cm]|S[table-format=2.2, table-column-width=0.95cm]|S[table-format=2.2, table-column-width=0.58cm]S[table-format=2.2, table-column-width=0.58cm]S[table-format=1.2, table-column-width=0.58cm]|S[table-format=2.2, table-column-width=0.42cm]S[table-format=2.2, table-column-width=0.42cm]S[table-format=2.2, table-column-width=0.42cm]@{}}
    \toprule
    \multirow{2}{*}{\textbf{Model}} & \multirow{2}{*}{\textbf{Params}} & \multicolumn{7}{c|}{\textbf{VG150 Dataset}} & \multicolumn{7}{c}{\textbf{PSG Dataset}} \\
    \cmidrule(lr){3-9} \cmidrule(lr){10-16}
    & & \textbf{Failure Rate} & \textbf{Recall} & \textbf{mRecall} & \textbf{zsRecall} & \textbf{Head} & \textbf{Body} & \textbf{Tail} & \textbf{Failure Rate} & \textbf{Recall} & \textbf{mRecall} & \textbf{zsRecall} & \textbf{Head} & \textbf{Body} & \textbf{Tail} \\
    \midrule
    
    \rowcolor{gray!15}
    \multicolumn{16}{l}{\textit{Zero-shot Performance}} \\
    \midrule
    Qwen2.5-VL & 3B & 76.95 & 0.46 & 0.34 & 0.12 & 0.46 & 0.31 & 0.13 & 51.05 & 3.69 & 3.69 & 0.00 & 2.66 & 4.15 & 2.42 \\
    Qwen2.5-VL & 7B & 24.89 & 3.40 & 2.38 & 0.78 & 3.09 & 3.26 & 1.98    & 8.69 & 8.78 & 11.02 & 0.00 & 6.82 & 10.45 & 7.25 \\
    Qwen2.5-VL & 32B & 9.95 & 8.40 & 5.66 & 1.68 & 8.20 & 6.13 & 4.96 & 5.67 & 15.69 & 17.38 & 0.00 & 14.03 & 18.60 & 13.90 \\
    \midrule
    
    \rowcolor{gray!15}
    \multicolumn{16}{l}{\textit{Fine-tuned Performance}} \\
    \midrule
    Qwen2.5-VL & 3B & 0.06  & 35.95 & 14.78 & 6.11 & 36.98 & 15.58 & 8.13 & 0.00 & 52.49 & 44.30 & 7.69 & 50.95 & 50.84 & 34.14 \\
    \bottomrule
  \end{tabular}
  \caption{Performance ($\%$) comparison of Qwen2.5-VL and our fine-tuned approach on the VG150 and PSG test sets.}
  \label{tab:zero-shot}
\end{table*}

\subsection{MLLM Zero-shot Performance in SGG}
\label{app:MLLM Zero-Shot performance in SGG}

We evaluate the zero-shot end-to-end SGG capabilities of Qwen2.5-VL (3B, 7B, 32B) using our CoT prompt. As shown in Table~\ref{tab:zero-shot}, the results demonstrate that MLLMs possess the untapped potential to handle dense relational reasoning in a streamlined, end-to-end fashion, bypassing the need for complex, multi-stage pipelines.

However, zero-shot performance remains subpar across all metrics, with Recall, mRecall, and zsRecall persisting at extremely low levels compared to our fine-tuned methods. Severe deficits across head, body, and tail categories indicate a fundamental failure to capture complex relational dependencies. While scaling the model size significantly enhances instruction following ability and reduces failure rates, it does not resolve the core bottleneck in relational reasoning. For instance, despite the 32B model achieving a low $9.95\%$ failure rate on VG150, its Recall of $8.40\%$ is strikingly inferior to the $35.95\%$ achieved by our fine-tuned approach. A similar disparity on PSG underscores the limitations of general-purpose MLLMs in dense relational reasoning under zero-shot settings, further validating the necessity of our proposed training pipeline.

\subsection{Qualitative Analysis on Relation Augmentation}
\label{app:Relation Augmentation Analysis}

To mitigate implausible relations generated by Qwen2.5-VL-32B, such as \textit{<hand.1, at, person.1>}, candidate relations are retained only if their maximum embedding cosine similarity to any ground-truth predicate exceeds a threshold. This refinement filters out $23.19\%$ and $13.83\%$ of initial outputs for VG150 and PSG, respectively, enhancing data quality.

As illustrated in Figure~\ref{fig:ra}, while the original datasets feature dense object annotations, they are often characterized by severe relation sparsity. Our augmentation strategy effectively recovers plausible inter-object associations, substantially enhancing the connectivity and completeness of the resulting scene graphs. For instance, it supplements missing spatial relations, such as \textit{<paper-merged.1, beside, tv.1>} and \textit{<mouse.2, on, table-merged.1>} for PSG samples, while clarifying possessive dependencies like \textit{<building.1, has, window.1>} and \textit{<sign.1, attached to, pole.1>} for VG150 samples. These relations are frequently omitted in the original annotations, and our method effectively mitigates this sparsity by generating more plausible relations.

\begin{figure*}[h]
  \centering
  \begin{subfigure}[b]{1.0\textwidth}
    \centering
    \includegraphics[width=\linewidth]{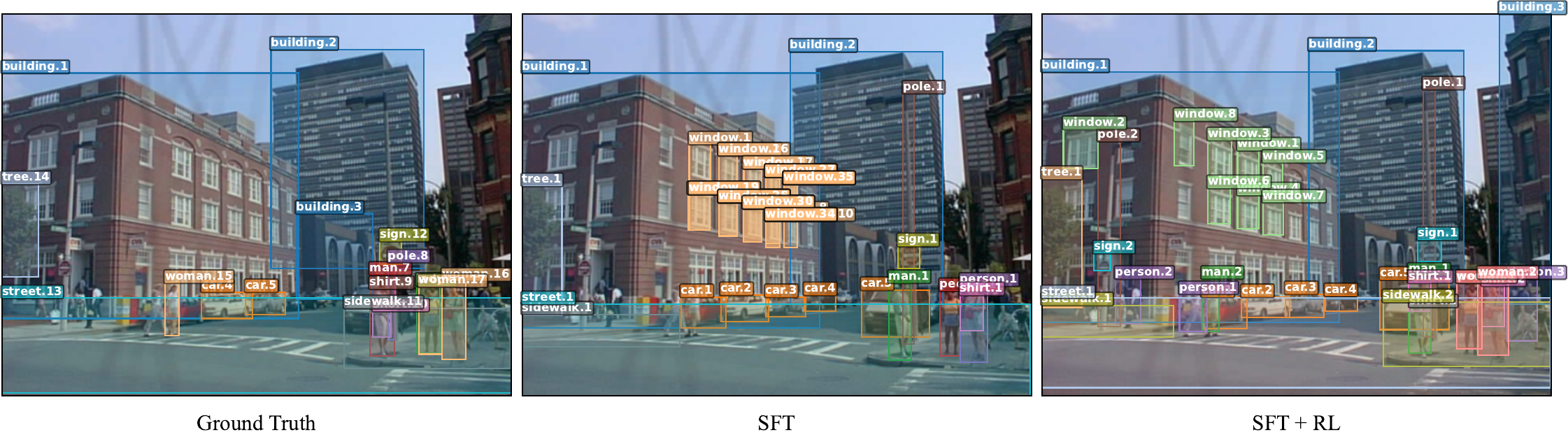}
    \caption{Qualitative visualization of object detection. Results are shown for \textbf{ground truth (left)}, the \textbf{SFT only output (middle)}, and \textbf{output after SFT followed by RL (right)}.}
    \label{fig:gt_sft_rl}
  \end{subfigure}
 \\[1em]
  \begin{subfigure}[b]{1.0\textwidth}
    \centering
    \includegraphics[width=\linewidth]{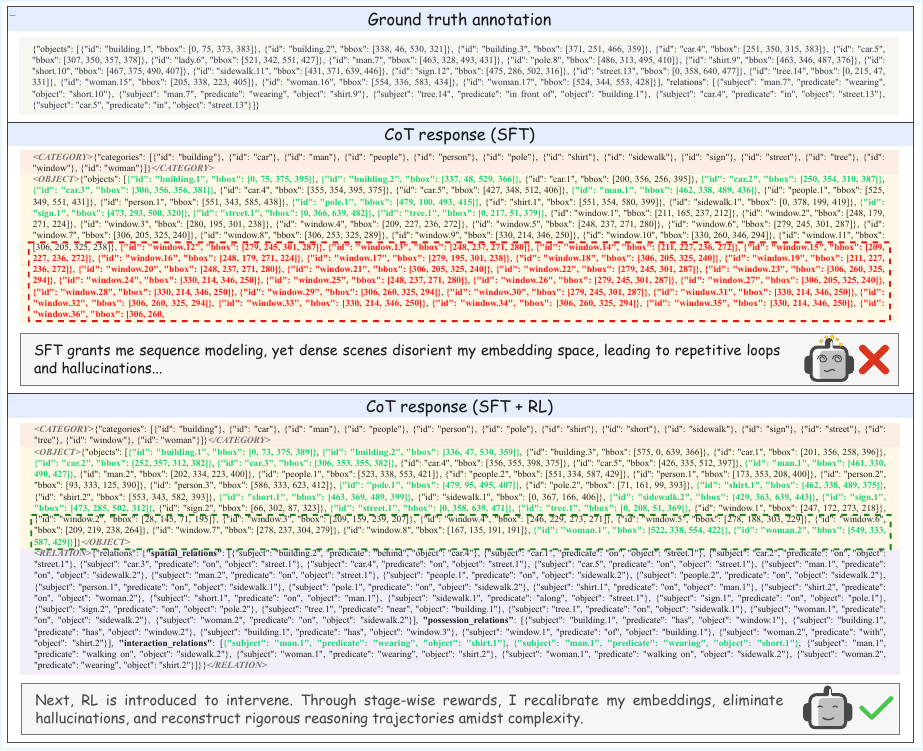}
    \caption{Comparison between SFT-only and SFT + RL outputs. Predictions are color-coded: \textcolor{green}{green} indicates alignment with ground truth, \textcolor{red}{red} denotes implausible repeated predictions.}
    \label{fig:sft_rl}
  \end{subfigure}
  \caption{Qualitative comparison of scene graph generation under different fine-tuning strategies. We visualize the ground truth, SFT-only outputs, and SFT + RL outputs. The comparisons highlight the performance gaps in both object detection (a) and relation recognition (b).}
  \label{fig:sft}
\end{figure*}

\begin{figure*}[h]
  \centering
  \begin{subfigure}[b]{1.0\textwidth}
    \centering
    \includegraphics[width=\linewidth]{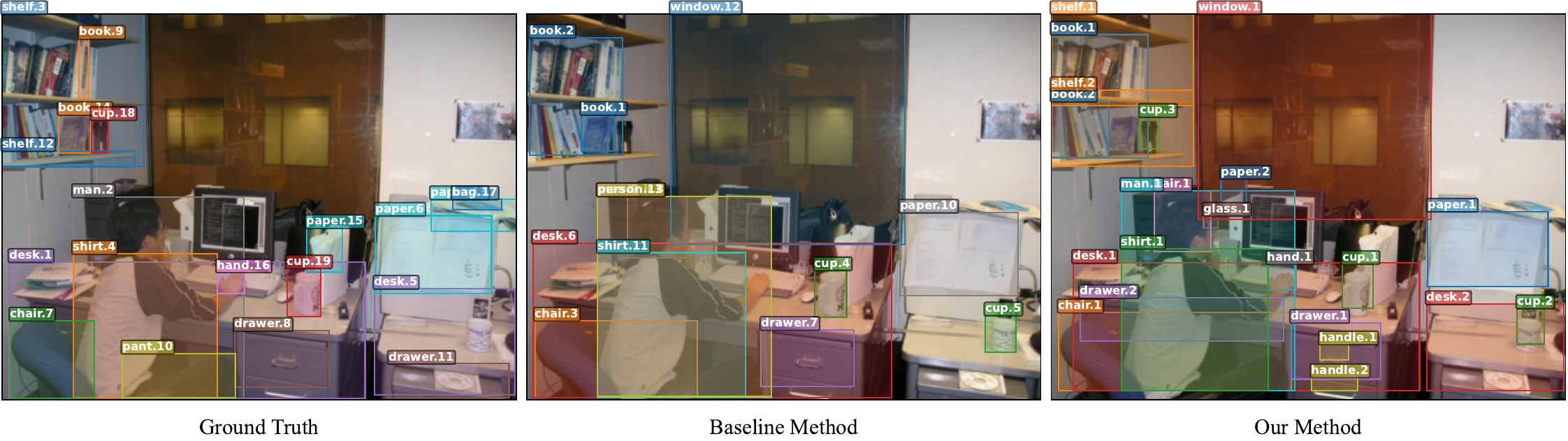}
    \caption{Qualitative visualization of object detection. Results are shown for \textbf{ground truth (left)}, the \textbf{baseline (middle)}, and \textbf{our method (right)}.}
    \label{fig:detection_scene1}
  \end{subfigure}
\\[1em]
  \begin{subfigure}[b]{1.0\textwidth}
    \centering
    \includegraphics[width=\linewidth]{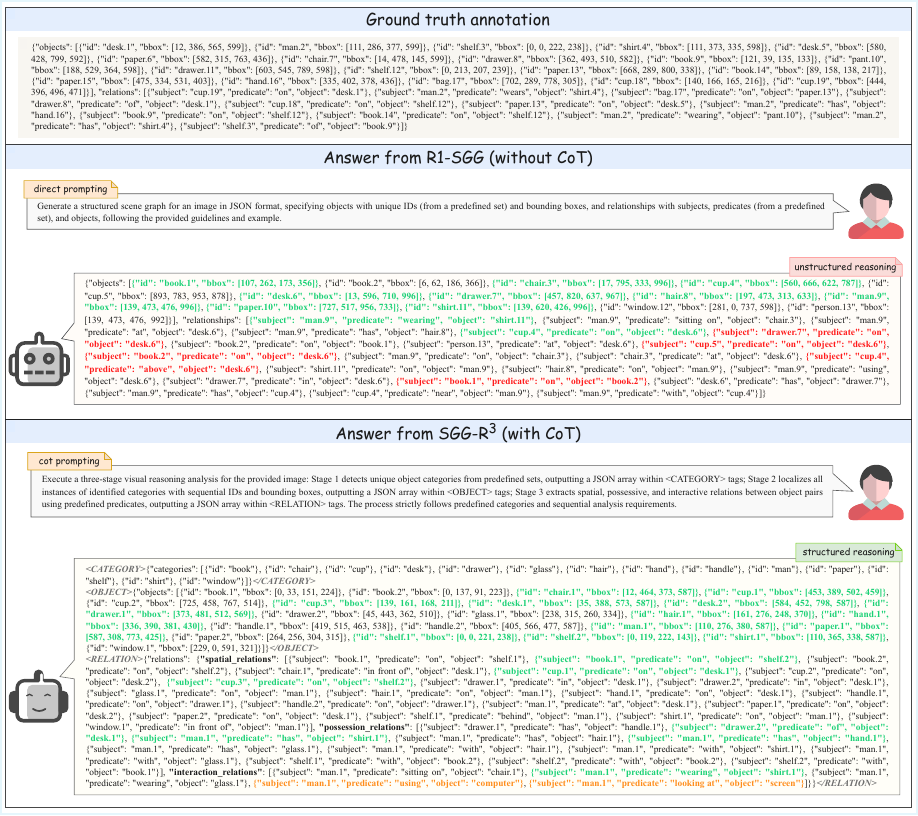}
    \caption{Qualitative visualization of reasoning trajectories. Object instances and relation triplets are color-coded: \textcolor{green}{green} indicates alignment with ground truth, \textcolor{red}{red} denotes implausible predictions, and \textcolor{orange}{orange} highlights plausible predictions for undetected objects.}
    \label{fig:detection_scene2}
  \end{subfigure}
  \caption{Qualitative comparison of scene graph generation between our method and the baseline. We visualize the ground truth, baseline reasoning outputs, and our method's CoT-based reasoning results. The comparisons highlight the performance gaps in both object detection (a) and relation recognition (b).}
  \label{fig:sggv}
\end{figure*}

\subsection{Qualitative Analysis of End-to-End SGG with SFT and RL}
\label{app:sft}

Comparative analysis in Figure~\ref{fig:sft} indicates that the SFT-only model is prone to hallucinations during dense object localization, exhibiting autoregressive degeneration such as redundant instances of \textit{window}. This behavior triggers sequence length limits, leading to premature truncation of outputs and underscoring the limitations of SFT. In contrast, RL optimization effectively eliminates these repetitive loops, enabling precise localization of eight \textit{window} instances. Notably, the outputs maintain strict cross-stage sequentiality: object localization adheres to the identified category order, while relation extraction follows the subject sequence. This structural consistency focuses the model's attention, ensuring complete and coherent reasoning, which significantly enhances recall performance.

\noindent \textbf{Object Detection Performance.} The output of the SFT model demonstrates incomplete object detection results with low recall. Compared to the SFT model outputs, it is evident that for objects such as \textit{sidewalk}, \textit{building}, \textit{pole}, and \textit{sign}, the RL model can detect a greater number of reasonable instances of these categories, which is beneficial for improving ground truth coverage. Furthermore, the detection accuracy of the RL model is also enhanced, with fine-grained objects such as \textit{sign}, \textit{shirt}, and \textit{short} all being well-detected. Despite the sparse annotation of objects in the original images, the model is capable of detecting objects like \textit{window} even when they are unlabeled, demonstrating strong generalization.

\noindent \textbf{Relation Recognition Performance.} In terms of relation output, the SFT output suffers from truncation, leading to failed relation recognition. In contrast, the RL output is complete and includes all three types of relations. It can be observed that the model strictly adheres to recognizing and outputting these three types of relations, identifying relationships between objects from spatial, possessive, and interactive perspectives. Even with the original sparse relation annotations, our two-stage fine-tuning framework enables the model to output numerous reasonable relations, thereby improving recall and exhibiting strong generalization.

\subsection{Qualitative Comparison of End-to-End SGG}
\label{app:sggv}

Qualitative comparisons in Figure~\ref{fig:sggv} demonstrate the efficacy of our method relative to the baseline method R1-SGG~\cite{chen2025compile}. Although R1-SGG produces scene graphs after two-stage post-training (SFT and RL), its outputs are notably shorter, resulting in lower overall recall of correct predictions. In contrast, our approach consistently generates structured three-stage outputs with reasoning trajectories that frequently surpass the ground truth in depth, leading to significantly higher overall recall.

\noindent \textbf{Object Detection Performance.} Regarding object detection, it is evident that the baseline method exhibits restricted recall and frequently overlooks common objects such as \textit{cup} and \textit{shelf}, resulting in a sparse representation of the scene. In contrast, our model identifies a comprehensive range of plausible objects, including \textit{handle}, \textit{glass}, and \textit{cup}, even when these instances are omitted from the original annotations, thereby significantly improving object detection recall. These results underscore the superior grounding capabilities of our framework.

\noindent \textbf{Relation Recognition Performance.} The baseline yields sparse, often semantically trivial or erroneous relations, including unreasonable ones (e.g., \textit{<cup.5, on, desk.6>} with misinterpreted referents) and incorrect predicates (e.g., \textit{<draw.7, on, desk.6>}, \textit{<book.2, on, desk.6>}). In contrast, our approach generates a higher density of valid triplets across three semantic levels-spatial (e.g., \textit{<drawer.1, in, desk.1>}), possessive (e.g., \textit{<drawer.1, has, handle.1>}), and complex interactions (e.g., \textit{<person.1, wearing, glasses.1>})-achieving superior ground-truth coverage and semantic depth, thereby mitigating relation bias. Moreover, the model detects plausible relations even without explicit localization (e.g., \textit{<person.1, using, computer>} when the \textit{computer} instance is undetected), demonstrating strong generalization.

Collectively, these results demonstrate that our three-stage reasoning framework effectively enhances the model's capacity to recognize fine-grained objects and capture intricate semantic dependencies, advancing the pursuit of robust scene understanding in real-world environments.

\end{document}